\documentclass{article}
\usepackage{data/template/ijcai2025/ijcai25}

\usepackage{times}
\usepackage{soul}
\usepackage{url}
\usepackage[hidelinks]{hyperref}
\usepackage[utf8]{inputenc}
\usepackage[small]{caption}
\usepackage{graphicx}
\usepackage{amsmath}
\usepackage{amsthm}
\usepackage{booktabs}
\usepackage{algorithm}
\usepackage{algorithmic}
\usepackage[switch]{lineno}

\usepackage[T1]{fontenc}    
\usepackage{hyperref}       

\usepackage{amsfonts}       
\usepackage{nicefrac}       
\usepackage{microtype}      

\usepackage{xcolor}         
\usepackage{bm}
\usepackage{multirow}
\usepackage{pifont}
\usepackage[export]{adjustbox}
\usepackage{subcaption}
\usepackage{xspace}
\usepackage{mdframed}
\usepackage[multiple]{footmisc}

\newcommand{\citet}[1]{\textit{et al.} \shortcite{#1}}
\newcommand{\citep}[1]{\cite{#1}}

\title{Text-to-Image Alignment in Denoising-Based Models through Step Selection}

\author{Paul Grimal, Hervé Le Borgne \& Olivier Ferret \\
   Université Paris-Saclay, \\ 
   CEA, List, \\
   F-91120, Palaiseau, France \\ 
\texttt{\{paul.grimal, herve.le-borgne, olivier.ferret \}@cea.fr}
}

\begin{document}
\pagestyle{plain}
\setcounter{page}{1}
\maketitle

\newcommand{\sizeimgqual}{0.48}
\newcommand{\sizecolqual}{0.23}
\newcommand{\sizefirstcolqual}{0.02}
\newcommand{\sepimage}{0pt}

\newcommand{\imagerow}[9]{
    \begin{minipage}[c]{\sizefirstcolqual\linewidth} 
        \raggedleft
        \rotatebox{90}{#1} 
    \end{minipage}
    \begin{minipage}[c]{\sizecolqual\linewidth}
        \centering
        \includegraphics[width=\sizeimgqual\linewidth]{#2} 
        \includegraphics[width=\sizeimgqual\linewidth]{#3} 
    \end{minipage}
    \hspace{\sepimage}
    \begin{minipage}[c]{\sizecolqual\linewidth}
        \centering
        \includegraphics[width=\sizeimgqual\linewidth]{#4}
        \includegraphics[width=\sizeimgqual\linewidth]{#5}
    \end{minipage}
    \hspace{\sepimage}
    \begin{minipage}[c]{\sizecolqual\linewidth}
        \centering
        \includegraphics[width=\sizeimgqual\linewidth]{#6}
        \includegraphics[width=\sizeimgqual\linewidth]{#7}
    \end{minipage}
    \hspace{\sepimage}
    \begin{minipage}[c]{\sizecolqual\linewidth}
        \centering
        \includegraphics[width=\sizeimgqual\linewidth]{#8}
        \includegraphics[width=\sizeimgqual\linewidth]{#9}
    \end{minipage}
    \hfill
}
\newcommand{\captionrow}[4]{
    \begin{minipage}[c]{\sizefirstcolqual\linewidth} 
        \raggedleft
        \rotatebox{90}{\space}
    \end{minipage}
    \begin{minipage}[c]{\sizecolqual\linewidth}
        \centering
        #1
    \end{minipage}
    \hspace{\sepimage}
    \begin{minipage}[c]{\sizecolqual\linewidth}
        \centering
        #2
    \end{minipage}
    \hspace{\sepimage}
    \begin{minipage}[c]{\sizecolqual\linewidth}
        \centering
        #3
    \end{minipage}
    \hspace{\sepimage}
    \begin{minipage}[c]{\sizecolqual\linewidth}
        \centering
        #4
    \end{minipage}
    \hfill
}

\newcommand{\sizeimgqualshort}{0.48}
\newcommand{\sizecolqualshort}{0.43}
\newcommand{\sizefirstcolqualshort}{0.02}
\newcommand{\sepimageshort}{0pt}

\newcommand{\imagerowshort}[5]{
    \begin{minipage}[c]{\sizefirstcolqualshort\linewidth} 
        \raggedleft
        \rotatebox{90}{#1} 
    \end{minipage}
    \begin{minipage}[c]{\sizecolqualshort\linewidth}
        \centering
        \includegraphics[width=\sizeimgqualshort\linewidth]{#2} 
        \includegraphics[width=\sizeimgqualshort\linewidth]{#3} 
    \end{minipage}
    \hspace{\sepimageshort}
    \begin{minipage}[c]{\sizecolqualshort\linewidth}
        \centering
        \includegraphics[width=\sizeimgqualshort\linewidth]{#4}
        \includegraphics[width=\sizeimgqualshort\linewidth]{#5}
    \end{minipage}
    \hfill
}
\newcommand{\captionrowshort}[2]{
    \begin{minipage}[c]{\sizefirstcolqualshort\linewidth} 
        \raggedleft
        \rotatebox{90}{\space}
    \end{minipage}
    \begin{minipage}[c]{\sizecolqualshort\linewidth}
        \centering
        #1
    \end{minipage}
    \hspace{\sepimageshort}
    \begin{minipage}[c]{\sizecolqualshort\linewidth}
        \centering
        #2
    \end{minipage}
    \hfill
}

\newcommand{\cmark}{\ding{51}}%
\newcommand{\xmark}{\ding{55}}%
\newcommand{\ie}{\textit{i.e.}\xspace}
\newcommand{\eg}{\textit{e.g.}\xspace}
\newcommand{\supmat}{Supplementary Material\xspace}
\newcommand{\gsng}{GSNg\xspace}
\newcommand{\iteref}{IterRef\xspace}
\newcommand{\hx}{\hat{x}_0}
\newcommand{\sdun}{SD~1.4\xspace}
\newcommand{\sdtrois}{SD~3\xspace}

\begin{abstract}
    Visual generative AI models often encounter challenges related to text-image alignment and reasoning limitations. This paper presents a novel method for selectively enhancing the signal at critical denoising steps, optimizing image generation based on input semantics. Our approach addresses the shortcomings of early-stage signal modifications, demonstrating that adjustments made at later stages yield superior results. We conduct extensive experiments to validate the effectiveness of our method in producing semantically aligned images on Diffusion and Flow Matching model, achieving state-of-the-art performance. Our results highlight the importance of a judicious choice of sampling stage to improve performance and overall image alignment. This work is an updated and extended version of the preprint originally published on OpenReview~\cite{grimal2025signal}. The code is available at \url{https://github.com/grimalPaul/gsn-factory}.
\end{abstract}

\section{Introduction}

Visual Generative AI Models usually rely on denoising processes such as Diffusion models~\cite{ho2020denoising} or Flow Matching~\cite{lipman2023flow}. They can be conditioned by a textual prompt to guide the inference, resulting in visually pleasing images~\cite{rombach2021highresolution,podell2023sdxl,ramesh2022hierarchical,saharia2022photorealistic}.
Although these models show impressive semantic and compositional capacities, even the best models still suffer from text-image alignment and reasoning limitations (\eg spatial, counting). Some works address these issues by improving the noisy captions in training datasets \cite{chen2023pixartalpha,chen2024pixartsigma,segalis2023picture} or improving the architecture \cite{Peebles2022DiT}, while others adopt an attention-based Generative Semantic Nursing (GSN) approach~\cite{chefer2023attendandexcite,rassin2023linguistic,guo2024initno} that avoids retraining the whole model by correcting it at inference or adding conditioning to better guide the generation.

\begin{figure}[h]
    \centering
    \input{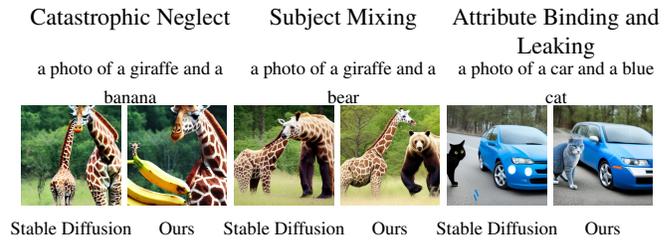}
    \caption{
        Samples generated by Stable Diffusion vs. Ours.}
    \label{fig:ti_pbm}
\end{figure}
Early research has identified several text-image alignment issues \cite{ramesh2022hierarchical,saharia2022photorealistic,chefer2023attendandexcite,feng2023trainingfreestructureddiffusionguidance}. They include \textit{catastrophic neglect}, where one or more elements of the prompt fail to be generated; \textit{subject mixing}, where distinct elements are improperly combined; \textit{attribute binding}, where attributes (\eg color) are incorrectly assigned to an entity; and \textit{attribute leaking}, where attributes are correctly bound to the specified elements but are erroneously applied to additional, unintended elements in the scene (\autoref{fig:ti_pbm}).

\begin{figure*}[h]
    \centering
    \includegraphics[width=0.7\linewidth]{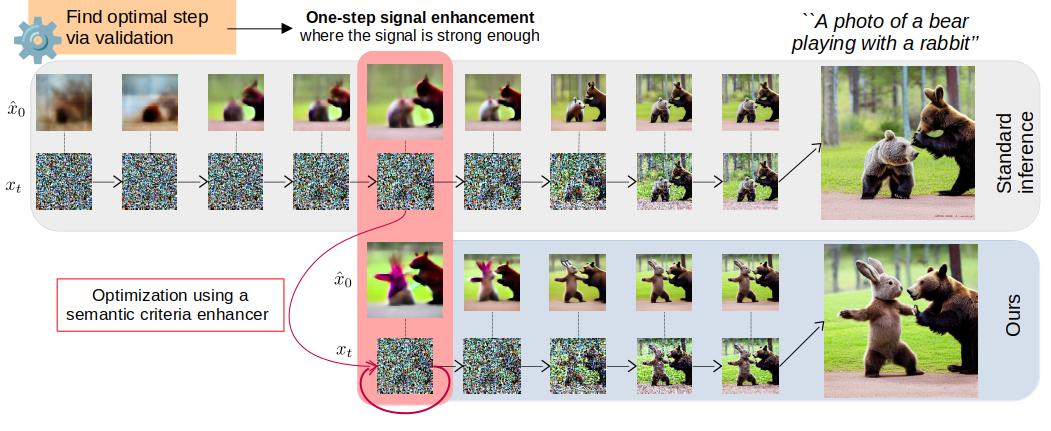}
    \caption{The diffusion process is paused at a key step (determined on a validation subset) to enhance the signal in the latent image. By amplifying the signal at this critical point, we ensure that the model can correctly construct the main components of the image, leading to a more accurate final result.}
    \label{fig:schem_method}
\end{figure*}

To improve generation, training-free methods  \cite{chefer2023attendandexcite,rassin2023linguistic,li2023divide,guo2024initno,Agarwal_2023_ICCV} have emerged. These methods leverage the text-image relationship in the model’s diffusion features to optimize the latent image that the diffusion model is denoising to adjust it. However, these approaches require testing and carefully selecting multiple sensitive hyperparameters (\eg choosing various diffusion steps to perform optimization or setting different loss thresholds to reach for each diffusion step), which can lead to potential failures during the optimization process. In addition, although multiple refinement steps are commonly employed along the diffusion path, the necessity for their repeated use and the reasoning behind their placement have been determined largely through experimental results, without clear explanation. We argue that a closer examination of the location of refinement steps would not only improve performance but also provide a better understanding of the optimal location of these steps.
To mitigate the risk of under/over optimization, InitNO~\cite{guo2024initno} optimizes multiple initial latent images solely at the first diffusion step, where latent images are pure Gaussian noise. However, the diffusion model’s reverse process reconstructs the signal gradually during image generation, making early-stage optimization less effective due to the weak signal at that point. As the signal becomes stronger in later diffusion steps, it provides more useful information for the refinement of the latent image. A deeper understanding of signal degradation dynamics can be used to improve generation capacity. In this work, we examine the impact of selecting the optimal steps to enhance the signal based on semantic content and demonstrate that carefully selecting these steps leads to substantial improvements in text-to-image alignment.

The main contributions of the paper are  (1) an extension of the Generative Semantic Nursing  to the architecture of the state-of-the-art Flow Matching model Stable Diffusion 3 and (2) a method for selectively enhancing the signal at a key step of the diffusion or flow matching process, optimizing image generation based on the input semantics. We demonstrate that early-stage signal modification is less effective and show that later adjustments lead to improved results.  We validate our approach through extensive experiments, demonstrating its effectiveness in producing semantically aligned images and achieving state-of-the-art results while also studying the placement of the refinement steps.

\section{Related Work}

\paragraph{New controls}

Li~\citet{li2023gligen} and Mou~\citet{mou2023t2i} introduce trainable modules to enable the addition of new conditions to the frozen models. Similarly, Zhang~\citet{zhang2023adding} incorporate a trainable copy of the model that can be conditioned on various control inputs, such as a drawing, a bounding box, or a depth map. Recent research focuses on conditioning models by working on the noisy latent image. SDEdit~\cite{meng2022sdedit} adds varying levels of noise to an image, balancing between fidelity to the original image and creative variation. Sun~\citet{sun2024spatialaware} create pseudo-guide images by placing objects on a background, adding noise, and then denoising them to maintain object placement during generation. Choi~\citet{Choi_2021_ICCV} inject down-sampled guide images during diffusion to create variations of the guided image.

FreeDoM~\cite{yu2023freedom} applies multiple updates to the latent image at various sampling steps and relies on an external model for guidance. In contrast, our method performs a single step of refinement using the model's inherent knowledge. This ensures better alignment and continuity in the generation process. External models optimizing blurry images may not understand the signal in the same way as the denoising model, leading to less precise results.
For example, a blurred image of three different entities makes it difficult to distinguish them, whereas in the semantic space of the model, the signal parts corresponding to each entity may be more distinct.

\paragraph{GSN}
GSN was introduced by Attend\&Excite~\cite{chefer2023attendandexcite}, aiming to optimize the latent image during inference to better consider semantic information without having to retrain models. The latent image $x_t$ at step $t$ is modified by applying gradient descent step w.r.t a loss $ \mathcal{L}$ on the extracted features produced by the model with the input $x_t$ : $x_{t'} \leftarrow x_t - \alpha_{t} \cdot \nabla_{x_t} \mathcal{L}$ ($\alpha_{t}$ the learning rate). Hence, it shifts the latent image to achieve the objective conceptualized by the loss function. Attend\&Excite considers the cross-attention features, which establish a link between image and text features, to ensure that the model adequately generates the subjects in the prompt. Building upon this approach, Syngen~\cite{rassin2023linguistic}, Divide and Bind~\cite{li2023divide}, InitNO~\cite{guo2024initno} and A-Star~\cite{Agarwal_2023_ICCV} design other loss functions to better enhance the alignment of the prompts while other works combine layout information to textual information to force objects placement~\cite{Chen_2024_WACV,Xie_2023_ICCV}.
The closest work to ours is InitNO, which performs a warm-up multi-round optimization on the initial latent image (initial noise). That is, they attempt to shift the initial latent image to reach a desired loss score, aiming to find an initial noise that will perform better during the generation process. The term ``multi-round'' applies because this process can take up to five rounds if the target loss score is not met, with a new initial latent image being resampled and optimized each time. In contrast, we argue that the optimization of the latent image is more effective at a later step than at the initial step. As the partial information of the latent image becomes progressively more accurate, it is beneficial to refine the information at a distant step, where the latent image is easier to distinguish from the noise, where the diffusion has a more accurate understanding of the signal in the latent image. In addition, our method is more efficient without the use of multi-round optimization.

\paragraph{Signal leak in diffusion models}

Lin~\citet{Lin_2024_WACV} reveal that Stable Diffusion 1.4 and some other diffusion models exhibit signal leakage, meaning the signal does not completely vanish even in the final steps of the forward process. Everaert~\citet{Everaert_2024_WACV} exploit this signal leakage to gain control over the generated images, biasing the generation towards desired styles, enhancing image variety, and influencing colors and brightness. Grimal~\citet{Grimal_2024_WACV} demonstrate that certain noises during inference perform better for generating multiple objects. We hypothesize that this performance arises from a signal in the initial noise, which is more consistent in making multiple objects appear. Based on the signal construction during the denoising, we identify the diffusion step where we can improve the signal and align it with the text.

\section{Methodology}

\subsection{Preliminary: Diffusion and Flow Matching}

Stable Diffusion~1.4 (\sdun)~\cite{rombach2021highresolution} and Stable Diffusion~3 (\sdtrois)~\cite{esser2024scalingrectifiedflowtransformers} are based on Diffusion Models~\cite{ho2020denoising} and Flow Matching~\cite{lipman2023flow}, respectively. Both rely on a forward process that gradually corrupts images over time $t$ and a reverse process that removes noise step by step. Although these frameworks are close, they differ in their core mechanism. Both learn to go from a distribution $p_{\text{source}}$ to $p_{\text{target}}$, which, in our case, is the manifold of the images.

\paragraph{Diffusion model}
The sampling process in diffusion models is stochastic. It involves a joint distribution between noisy samples at different time steps $t$ and $t'$. At each step $t'$, the model estimates the expected value of the previous sample $\mathbb{E}[x_{t'} \vert x_t]$ and resamples $x_{t'}$ from this estimated distribution. This process is repeated iteratively until the final target distribution is reached. Equivalently, the model can be trained to predict the noise added at each step, which is captured by the following loss function, where $\epsilon_\theta$ is the model:
\begin{equation}
    \mathcal{L} = \underset{x, \epsilon \sim \mathcal{N}(0, I), t}{\mathbb{E}} \Big[ \|\epsilon - \epsilon_\theta(x_t, t)\|^2 \Big].
\end{equation}

\paragraph{Flow Matching}
The sampling process is deterministic. This method defines a path from the source distribution to the target distribution. The model learns a velocity field $v_t^{[\text{source}, \text{target}]}(x_t)$ that transports data from $p_{\text{source}}$ to $p_{\text{target}}$. Learning this field involves predicting $\mathbb{E}[v_t^{[\text{source}, \text{target}]}(x_t) \vert x_t]$ at each step. During training, the velocity field is known; for instance, we can choose a linear flow and have $v_t^{[\text{source}, \text{target}]} = x_0 - \epsilon$ with $\epsilon \sim p_{\text{source}}$ and $x_0 \sim p_{\text{target}}$, transporting $\epsilon$ linearly to $x_0$. The model $v_\theta(x_t, t)$ is learnt by minimizing:
\begin{equation}
    \mathcal{L} = \underset{x, \epsilon \sim \mathcal{N}(0, I), t}{\mathbb{E}} \Big[ \| v_t^{[\text{source}, \text{target}]}(x_t) -  v_\theta(x_t,t)  \|^2 \Big]
\end{equation}

\paragraph{Signal degradation by noise} Here, we consider the diffusion setup of \sdun and the linear flow matching from \sdtrois. In both cases, the initial distribution is Gaussian and the target distribution is the set of images. At time step $t$, data $x_t$ result from adding noise with a predefined schedule:
\begin{equation}
    x_t = a_t x_0 + b_t \epsilon,\quad \epsilon \sim \mathcal{N}(0, I),
    \label{eq:interpolation_noise}
\end{equation}
where $a_t$ and $b_t$ are determined by the noise scheduler. This can be viewed as interpolating between the original signal $x_0$ and noise $\epsilon$. Increasing $b_t$ makes the signal harder to distinguish from the noise, thus controlling the degradation level.

\paragraph{Text-conditioning} To condition the generation with text, several authors~\cite{rombach2021highresolution,podell2023sdxl,saharia2022photorealistic,chen2023pixartalpha,balaji2023ediffi} adopt a cross-attention mechanism consisting of using the embedding of a prompt $p$ from a frozen textual encoder $\tau(\cdot)$ like T5~\cite{raffel2020exploring} or CLIP~\cite{radford2021learning}. The textual encoder generates an embedding of $N$ tokens, which the model utilizes across different cross-attention layers. Within these layers, a linear projection is applied to the intermediate features $Q$ and the text embedding $K$. Attention maps are then computed as $A = \text{softmax}(QK^T/\sqrt{d})$. These attention maps can be reshaped into $\mathbb{R}^{h \times w \times N}$, where $h$ and $w$ represent the dimensions of the attention maps in the cross-attention layer, and $N$ denotes the sequence length of the prompt embedding. As demonstrated by \cite{hertz2022prompttopromptimageeditingcross,tang2023daam}, cross-attention maps reveal meaningful semantic relationships between the spatial layout and corresponding words, which can be utilized for visualization and control. The models are thus conditioned by $(x_t, t, \tau(p))$.

\paragraph{Computational efficiency} To reduce the computational cost of diffusion, Rombach~\citet{rombach2021highresolution} developed a Latent Diffusion Model (LDM) that operates within a smaller perceptual latent space. This model generates an initial latent noise $z_T$, denoises it iteratively to obtain $z_0$, and then projects the latent image into pixel space to produce the final image $x_0$. Although our experiments use an LDM, our approach is equally applicable in pixel space. For clarity, we will describe the method using $x_t$, even though our experiments are conducted in the latent space. During inference, we can generate an image without following the full training  steps by using a sampling scheduler that discretizes the denoising process into a reduced number of steps. For example, with the DDPM scheduler with \sdun and 50 sampling steps, the first sampling step 0 corresponds to step 981 of the original diffusion process, significantly reducing the number of steps required while maintaining generation quality.

\paragraph{GSN} To improve the process, recent approaches adopt two processes that can be combined but have different purposes. First, they adopt \textit{GSN guidance} (\gsng) such that the latent image $x_t$ at step $t$ is shifted by applying a (unique) gradient descent step w.r.t a loss $ \mathcal{L}$ that favors the alignment with the prompt, thus $x_t$: $x_{t'} \leftarrow x_t - \alpha_{t} \cdot \nabla_{x_t} \mathcal{L}$, with $\alpha_{t}$, the learning rate. Second, the process can be repeated at each of some predefined sampling steps $t_1\dots t_k$ until either $\mathcal{L}$ reaches a sufficient threshold or a specified maximum number of shifts has been made. This process is called \textit{iterative refinement} (\iteref) step. We argue that choosing carefully the step at which \iteref is performed allows us to do it once only, without needing to compare to a threshold, thus reducing the number of hyperparameters to set while leading to better results.

\subsection{Choosing the Right \iteref Steps to Enhance the Content}

\begin{figure*}[t]
    \centering
    \includegraphics[width=0.8\linewidth]{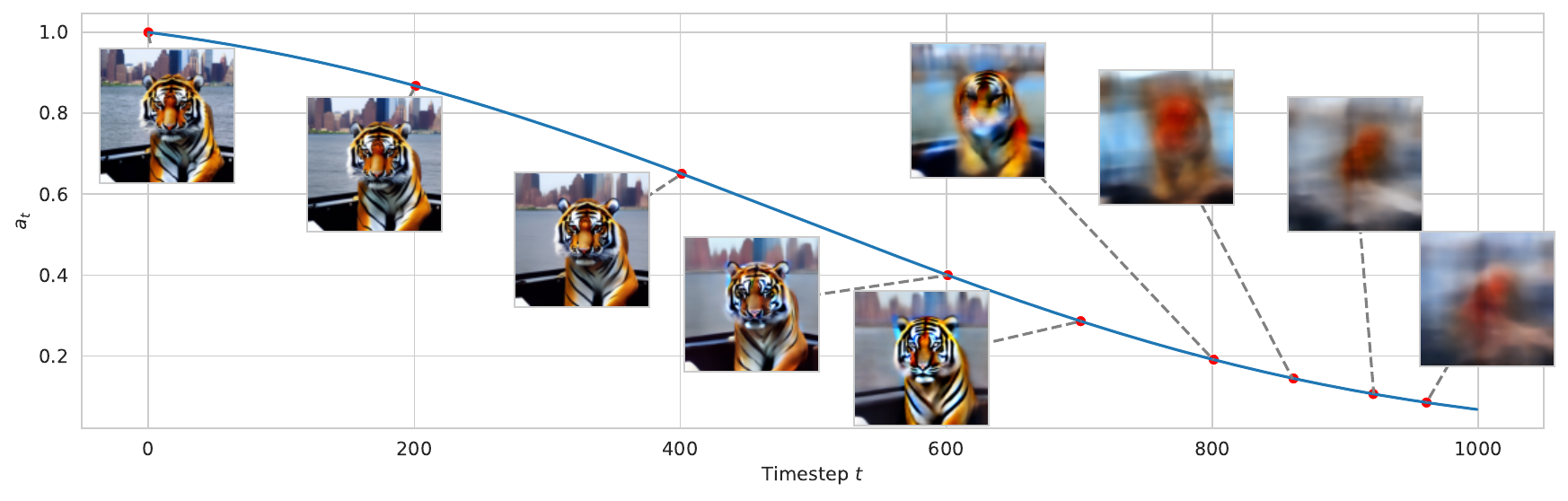}
    \caption{Value of $a_t$ as a function of the timestep $t$ ($t=0$ for the target distribution and $t=1000$ for the Gaussian). The estimated $\hx$ during the generation of ``a photo of a tiger on a boat arriving in new york'' at various steps is displayed. A coarse-to-fine generation is observed; as the denoising process progresses, the scene becomes increasingly distinguishable. Generated with Stable Diffusion 1.4.}
    \label{fig:snr_and_generation}
\end{figure*}

Our method focuses on selecting the appropriate denoising steps to enhance the signal within the noise, thereby generating more faithful final images.
Previous studies have shown the coarse-to-fine behavior of diffusion models \cite{park2023understanding}. In the following, we discuss how the signal is degraded and subsequently recovered by a diffusion model. The same conclusions can be drawn for Flow Matching, as the signal degradation follows the same process described in \autoref{eq:interpolation_noise}.

During the reverse process, the model reconstructs the low-frequency structure of the image first, before progressively refining the fine details towards the end. This behavior can be understood from the interpolation in \autoref{eq:interpolation_noise}, where, as the forward process progresses, the signal $x_0$ diminishes while the noise $\epsilon$ increases. Importantly, at each diffusion step, we can estimate the final image and obtain an approximation of the underlying signal.

Given any $x_t$ at a particular diffusion step, the final image $x_0$ can be estimated as:
\begin{equation}
    \hat{x}_0 =(x_t - b_t \epsilon_\theta (x_t))/a_t
\end{equation}

In Figure \ref{fig:snr_and_generation}, we visualize the estimated signal during the diffusion process, alongside the value of $a_t$. As the process progresses, the signal becomes more defined, allowing the general structure of the final image to emerge even in the early stages. The degradation and reconstruction of the signal $x_0$ are controlled by the noise scheduler. Previous studies \cite{Choi_2022_CVPR,chen2023importance} have emphasized the importance of carefully selecting the noise schedule to allocate sufficient time for the model to construct the main content of the image. This ensures that the model has ample opportunity to build the scene accurately. In the context of semantic image generation, this explains why attention to the text prompt is stronger at earlier noise levels when the core elements of the image are still being formed~\cite{balaji2023ediffi,park2023understanding}. At later stages, text input has less influence, the model focusing on refining details while the general spatial structure remains the same.

Our method leverages this understanding by enhancing the signal at the critical timesteps: neither too early, when the signal is weak, nor too late, when the scene is already defined. This ensures that the signal remains sufficiently strong throughout the reverse process, guiding the model to semantically construct the final image accurately. By carefully choosing the step, we can amplify the signal in the latent image, allowing for better semantic alignment with the text prompt. To select the best-performing steps automatically, we propose a validation method to test multiple steps on an evaluation metric (see \ref{sec:exp_settings}). Our approach is summarized in \autoref{fig:schem_method}.

Since the latent space of diffusion models inherently lacks semantic meaning \cite{kwon2023diffusionmodelssemanticlatent,park2023understanding}, making it unsuitable for direct manipulation to control the generated results, we rely on the model's ability to interpret the latent representation to assign semantic relevance and we use a single \iteref step to enhance the signal and ensure faithful alignment between text and image. In other words, we modify the signal interpreted by the model to enhance its quality, ensuring that the model receives an appropriate signal for a correct generation. Additionally, our only-one \iteref step approach is versatile and can be integrated with methods like \gsng for further improvement in image generation.

\subsection{Enhancing the Signal According to the Text-to-image Alignment Task}

Considering a prompt $p$ with a list of subject tokens $\mathcal{S} = \{s_1, \dots, s_k\}$, we extract attention features for each subject. For \sdun we follow \cite{chefer2023attendandexcite} to obtain one attention maps $A^s$ per subject $s$. For \sdtrois, the architecture integrates tranformers blocks called MM-DiT, where the latent image $x_t$ is patchified and processed alongside T5 and CLIP embeddings through attention mechanism. This mechanism can be seen as a combination of self-attention and cross-attention, making it challenging to extract meaningful attention maps. To address this, we isolate and refine the attention maps corresponding to CLIP and T5, average them across entities, and apply the GSN criterion for alignment. Further details are provided in the \supmat.
To ensure attention for each subject token, we consider:
\begin{equation}
    \mathcal{L}_{CN} = \underset{s \in S}{\max} (1 - \underset{i,j}{\max}(A^{s}_{i,j}))
\end{equation}
proposed by Chefer~\citet{chefer2023attendandexcite}, where $A^s_{i,j}$ represents the cross-attention value at position $i,j$ for the subject token $s$. It encourages the token with minimal activation to be more excited. Additionally, we implement an Intersection Over Union (IoU) loss, already used in \cite{Agarwal_2023_ICCV}, to mitigate catastrophic mixing by fostering subject separation. For all combinations of subject token pairs $\mathcal{C}$, the loss is defined as:
\begin{equation}
    \mathcal{L}_{\text{IoU}} =  \frac{1}{\vert \mathcal{C}\vert}\underset{\forall (m, n) \in \mathcal{C}}{\sum} \left( \frac{\underset{i,j}{\sum} \min(A^m_{i,j}, A^n_{i,j})}{\underset{i,j}{\sum} (A^m_{i,j} +A^n_{i,j} )} \right)
\end{equation}
where $A^s_{i,j}$ denotes the cross-attention value at position $i,j$ for subject token $s$. Finally, our loss is defined as $\mathcal{L} = \mathcal{L}_{CN} + \mathcal{L}_{\text{IoU}}$, which is minimized during 50 shifting steps of the latent image $x_t$ with the Adam optimizer \cite{kingma2017adammethodstochasticoptimization} and a learning rate of $1\times10^{-2}$. These hyperparameters are fixed according to previous studies for a fair comparison.

\section{Empirical Analysis and Results}

\subsection{Experimental Settings}\label{sec:exp_settings}
\paragraph{Implementations}

We use \sdun as all hyperparameters methods are based on this model. Images are generated using the DDPM Scheduler with 50 inference steps, on an Nvidia A100 80GB in Float 32 precision, with a Classifier-Free Guidance \cite{ho2022classifierfree} of 7.5. We compare our approach against other inference methods that rely solely on the internal knowledge of the model, including the standard inference of Stable Diffusion, Attend\&Excite, Divide\&Bind \cite{li2023divide}, InitNO, and Syngen. We exclude A-Star due to a lack of an official implementation and because InitNO reports superior results.  The authors of InitNO propose to couple their methods with \gsng and \iteref steps, which we refer to as InitNO+. We refer to our method as Ours and its variant incorporating the \gsng from Syngen as Ours+, where the \gsng is applied after the iterative refinement step. We also compare the results of \sdtrois with and without our approach. See \autoref{tab:method_gsn_presentation} for a summary of the methods and the \supmat for further details.

\begin{table}[tb]
    \centering
    \resizebox{1\linewidth}{!}{%
        \newcommand{\stepsamplinggsn}{2}
\newcommand{\stepours}{8}
\newcommand{\sizemultiround}{\normalsize}

\begin{tabular}{llllll}
    \toprule
    \multirow[c]{1}{*}{\textbf{Methods}} & \textbf{\iteref}   & \textbf{\iteref}                                     & \textbf{\iteref} & \multirow[c]{1}{*}{\textbf{\gsng}} & \textbf{Max Gradient }    \\
                                         & Which Step         & Reach Threshold                                      & Max Shift        &                                    & \textbf{Updates of} $x_t$ \\
    \midrule
    Syngen                               & \o                 & \o                                                   & \o               & 25 first steps                     & $25$                      \\
    \midrule
    Attend\&Excite                       & 0 10 20            & \cmark                                               & 20               & 25 first steps                     & $85$                      \\
    \midrule
    Divide\&Bind                         & 0 10 20            & \cmark                                               & 50               & 25 first steps                     & $175$                     \\
    \midrule
    InitNO                               & 0                  & \cmark {\sizemultiround up to 4 restart if it fails} & 50               & \o                                 & $90$                      \\
    \midrule
    \multirow[l]{2}{*}{InitNO+}          & 0                  & \cmark {\sizemultiround up to 4 restart if it fails} & 50               & \multirow[l]{2}{*}{25 first steps} & \multirow[l]{2}{*}{$315$} \\
                                         & 10 20              & \cmark                                               & 20               &                                    &                           \\
    \midrule
    Ours                                 & $\stepours$        & \o                                                   & 50               & \o                                 & $50$                      \\
    \midrule
    Ours+                                & $\stepsamplinggsn$ & \o                                                   & 50               & from 3 to 25                       & $73$                      \\
    \bottomrule
\end{tabular}
    }
    \caption{Overview of methods. Steps are given in terms of sampling scheduler. \textit{Max Shift} indicates the maximum predefined shifts applied if no threshold is met or if no threshold is used. \textit{Max Gradient Updates} refers to the maximum number of times the latent image is updated during the generation.}
    \label{tab:method_gsn_presentation}
\end{table}
\paragraph{Evaluation metrics}

Following previous GSN approaches, we employ the CLIP-based metrics proposed by \cite{chefer2023attendandexcite}, referred to in this paper as the Similarity Score, which includes Full Prompt Similarity, Minimum Object Similarity, and Text-Text Similarity. In contrast to previous methods, we also compute the CLIP Score~\cite{radford2021learning} to measure the average alignment between text and image embeddings. However, caution is necessary when using these CLIP-based metrics, as they often struggle with relational understanding, can misassociate objects with their attributes, and exhibit a significant lack of order sensitivity~\cite{yuksekgonul2023when}. We thus report the TIAM score~\cite{Grimal_2024_WACV}, a metric closely correlated with human judgment that reflects the proportion of correctly generated images. For each prompt, we generated multiple images, which were automatically evaluated to ensure the correct appearance of the requested entities and, where applicable, their attributes such as color. We also report the LAION’s aesthetic predictor~\cite{schuhmann2022laion5bopenlargescaledataset} on a scale from 1 to 10. Finally, we conducted a user study to complement the evaluation. Further details on the metrics are in the \supmat.

\paragraph{Datasets}

Following the recommended sampling method of TIAM~\cite{Grimal_2024_WACV}, we generated prompts for all possible combinations of two and three subject entities using 24 COCO labels~\cite{lin2015microsoft} and possibly colors. For each dataset, 300 prompts were sampled and 16 images per prompt were generated using the same 16 seeds to create the test set. Additionally, we created four validation datasets by sampling 10 prompts, distinct from the 300 test prompts, to determine the optimal \iteref step. The four datasets consist of two entities, two colored entities, three entities, and three colored entities.

\paragraph{Optimal \iteref step selection}

\begin{figure*}[]
    \centering
    \includegraphics[width=0.4\linewidth]{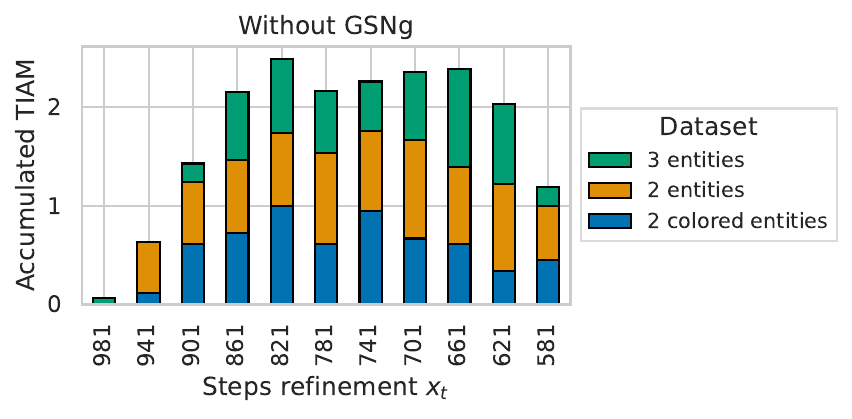}
    \includegraphics[width=0.4\linewidth]{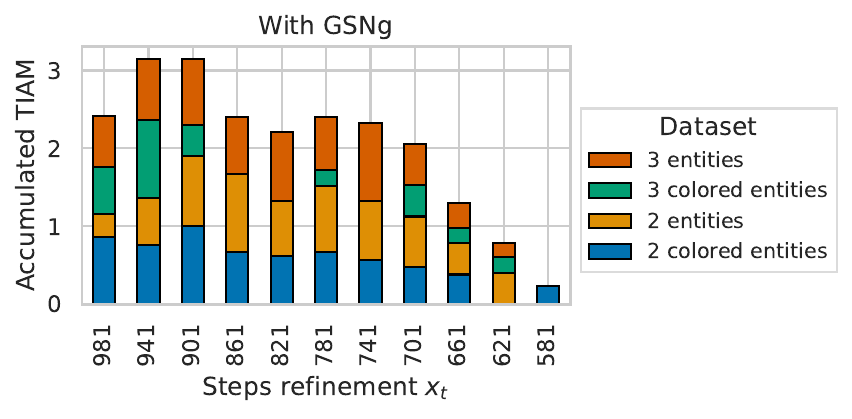}
    \caption{Accumulated TIAM scores without (left) and with (right) \gsng. The dataset with three colored entities is excluded on the left due to its low scores. Steps 821 and 941 are identified as optimal.}
    \label{fig:prompt_val_cumul_2_part}
\end{figure*}

We evaluate 11 sampling steps, spaced every two steps (\ie 0, 2, 4,\dots, 24) out of the 50 sampling steps for \sdun.
We focus on the first 25 steps, as prior research shows limited benefit beyond this point \cite{chefer2023attendandexcite}. For each validation dataset, we generate 16 images per prompt using the same 16 seeds and compute the TIAM score. We standardize the scores using a min-max scaler for each dataset and present the \textit{accumulated standardized TIAM} across the \iteref step for Ours and Ours+ in \autoref{fig:prompt_val_cumul_2_part}.
Based on the scores, we find that step 821 (sampling step 8) is optimal without \gsng, while step 941 (sampling step 2) produces better results with \gsng. This difference can be explained by the need for changes to occur later in the process without \gsng, ensuring that the adjusted signal is strong enough to persist through random sampling. In contrast, \gsng enables continuous signal refinement, allowing for corrections even at later stages. Moreover, we calculate the aesthetic score and observe no degradation whatever the \iteref step chosen, confirming the choice (values available in the \supmat). We will use these selected steps in subsequent experiments. Our validation method is computationally efficient, requiring only 10 prompts with a limited number of samples to determine the optimal \iteref step. We applied the same approach to select the best step for \sdtrois (details are in \supmat).

\subsection{Quantitative Results}

\paragraph{TIAM}

\begin{table*}[tb]
    \centering
    \begin{minipage}[t]{0.48\linewidth}
        \centering
        \resizebox{1\linewidth}{!}{
            \begin{tabular}{rcclcccc}
\toprule
& \multirow[c]{2}{*}{\rotatebox{0}{\shortstack{\iteref}}} &\multirow[c]{2}{*}{\rotatebox{0}{\shortstack{\gsng}}} & \multirow[c]{2}{*}{\rotatebox{0}{Methods}} & \multicolumn{2}{c}{w/o colors} & \multicolumn{2}{c}{with colors} \\
&  &  &  & 2 entities & 3 entities & 2 entities & 3 entities \\
\midrule
\multirow[c]{7}{*}{\rotatebox{90}{\textbf{SD 1.4}}} & 0 & \xmark & Stable Diffusion & $45.4_{32.2/5.5}$ & $8.4_{33.5/5.5}$ & $3.9_{34.6/5.4}$ & $0.1_{34.5/5.4}$ \\
\cmidrule(lr){2-8}
& \multirow[c]{2}{*}{\rotatebox{0}{1}} & \multirow[c]{2}{*}{\rotatebox{0}{\xmark}} & InitNO & $62.1_{33.1/5.5}$ & $14.2_{34.3/5.4}$ & $7.2_{35.7/5.4}$ & $0.2_{35.5/5.3}$ \\
&  &  & Ours & $65.8_{33.7/5.5}$ & $23.1_{35.4/5.5}$ & $8.7_{36.4/5.4}$ & $0.4_{36.3/5.4}$ \\
\cmidrule(lr){2-8}
& \multirow[c]{3}{*}{\rotatebox{0}{3}} & \multirow[c]{3}{*}{\rotatebox{0}{\cmark}} & Divide\&Bind & $69.9_{33.7/5.5}$ & $33.6_{35.9/5.4}$ & $11.3_{36.1/5.4}$ & $0.5_{36.1/5.3}$ \\
&  &  & Attend\&Excite & $71.4_{34.0/5.5}$ & $32.0_{35.9/5.4}$ & $10.5_{36.9/5.4}$ & $0.6_{36.9/5.3}$ \\
&  &  & InitNO+ & $75.0_{34.1/5.5}$ & $34.2_{36.0/5.4}$ & $11.9_{37.1/5.4}$ & $1.0_{37.3/5.3}$ \\
\cmidrule(lr){2-8}
& 0 & \cmark & Syngen & $\underline{78.5}_{34.1/5.4}$ & $\underline{39.2}_{36.5/5.4}$ & $\underline{20.4}_{37.1/5.3}$ & $\underline{2.4}_{36.8/5.3}$ \\
\cmidrule(lr){2-8}
& \multirow[c]{2}{*}{\rotatebox{0}{1}} & \multirow[c]{2}{*}{\rotatebox{0}{\cmark}} & Syngen+ & $75.8_{33.8/5.3}$ & $36.2_{36.2/5.4}$ & $20.1_{37.1/5.3}$ & $1.9_{36.9/5.3}$ \\
&  &  & Ours+ & $\textbf{81.1}_{34.2/5.4}$ & $\textbf{45.8}_{36.7/5.4}$ & $\textbf{20.5}_{37.1/5.3}$ & $\textbf{2.8}_{37.1/5.3}$ \\
\midrule
\multirow[c]{2}{*}{\rotatebox{90}{\textbf{SD 3}}} & 0 & \xmark & Stable Diffusion &${82.8}_{34.8/5.5}$ & ${63.4}_{37.9/5.5}$    & $\textbf{27.3}_{38.2/5.4}$ & ${9.69}_{39.4/5.3}$ \\
\cmidrule(lr){2-8}
 & 1 & \xmark & Ours &  $\textbf{84.5}_{34.9/5.6}$  & $\textbf{70.7}_{38.2/5.6}$ &  ${24.2}_{38.1/5.4}$ &  $\textbf{9.71}_{39.6/5.4}$ \\
\bottomrule
\end{tabular}

        }
        \caption{TIAM performance for prompts containing two or three entities, with and without color specifiers. The subscripts refer to {CLIP/aesthetic scores}. Best values are in bold, with second-best underlined for \sdun. For \sdtrois, only the best values are in bold.}
        \label{tab:tiam_score_qual}
    \end{minipage}
    \hfill
    \begin{minipage}[t]{0.48\linewidth}
        \centering
        \resizebox{1\linewidth}{!}{
            \begin{tabular}{rccllll}
\toprule
& \iteref &\gsng &Methods & Full Prompt & Minimum Object & Text-Text \\
\midrule
\multirow[c]{7}{*}{\rotatebox{90}{\textbf{SD 1.4}}} & 0 & \xmark & Stable Diffusion & 0.3313 & 0.2400 & 0.7682  \\
\cmidrule(lr){2-7}
& \multirow[c]{2}{*}{1} & \multirow[c]{2}{*}{\xmark} & InitNo & 0.3411 & 0.2512 & 0.7901 \\
&  &  & Ours & 0.3470 & 0.2564 & 0.7979 \\
\cmidrule(lr){2-7}
& \multirow[c]{3}{*}{3} & \multirow[c]{3}{*}{\cmark} & Divide\&Bind & 0.3468 & 0.2597  & 0.8065 \\
&  &  & Attend\&Excite & 0.3509 & 0.2634 & 0.8032 \\
& &  & InitNO+ & \underline{0.3520} & 0.2638 & 0.8076 \\
\cmidrule(lr){2-7}
& 0 & \cmark & Syngen & 0.3518 & \underline{0.2640} & \underline{0.8122} \\
\cmidrule(lr){2-7} 
& 1 & \cmark & Ours+ & \textbf{0.3522} & \textbf{0.2643} & \textbf{0.8133} \\
\midrule
\multirow[c]{2}{*}{\rotatebox{90}{\textbf{SD 3}}} & 0 & \xmark & Stable Diffusion & 0.3529 & 0.2616 & 0.8181 \\
\cmidrule(lr){2-7}
& 1 & \xmark & Ours & \textbf{0.3535} & \textbf{0.2619} & \textbf{0.8190}\\
\bottomrule
\end{tabular}

        }
        \caption{Similarity scores based on \protect\cite{chefer2023attendandexcite} for two entities. Best values are in bold, with second-best underlined for \sdun. For \sdtrois, only the best values are in bold.}
        \label{tab:its_2_objects_wo}
    \end{minipage}
\end{table*}

We present in \autoref{tab:tiam_score_qual} the TIAM, CLIP and aesthetics scores of our method and other approaches. With \sdun, our method outperforms InitNO across all configurations in both TIAM and CLIP scores, using a single \iteref step without \gsng. This indicates that a single \iteref step is more effective when the signal is stronger than at the initial diffusion step, as expected by our approach. When combined with \gsng, we surpass all other methods in terms of TIAM scores, showing that \gsng leads to better results with our carefully chosen \iteref step. We achieve superior CLIP scores in all configurations, except for the three colored entities, where TIAM alignment scores are generally very low across all methods. For a fair comparison, we tried to add an \iteref step for the Syngen approach, referred to as Syngen+, but obtained an even lower score. More details are in the \supmat. With \sdtrois, our method mitigates catastrophic neglect, showing improved TIAM and CLIP scores for two and three entities. We note a slight decrease in performance for two colored entities but nearly identical TIAM scores with a better CLIP score for three colored entities.

\paragraph{Similarity score}
\label{sec:similarity_score}

We present the scores for the dataset with two entities in \autoref{tab:its_2_objects_wo}. For \sdun, without \gsng our method consistently outperforms InitNO, confirming the importance of carefully selecting the diffusion step to perform the \iteref steps. With \gsng, we surpass all competing methods. While we achieve slightly better performance on datasets with two and three entities, Ours+ is marginally lower for datasets that include color specifications. This may be attributed to the limitations of CLIP-based metrics in capturing precise syntactic relations \cite{yuksekgonul2023when}. Ours outperforms \sdtrois on all datasets, with a minor drop in one metric for two colored entities. Results for the other datasets and further discussion on the limits of this score are in the \supmat.

\paragraph{User study}

\begin{table}[!ht]
    \centering
    \resizebox{0.8\linewidth}{!}{
        \begin{tabular}{lll|lll}
    \toprule
       & \multicolumn{2}{c|}{w/o \gsng} & \multicolumn{3}{c}{w \gsng}                                      \\
       & Ours                           & InitNO                      & Ours+           & Syngen & InitNO+ \\
    \midrule
    \% & \textbf{43.1\%}                & 36.9\%                      & \textbf{57.4\%} & 51.9\% & 43.3\%  \\
    \bottomrule
\end{tabular}
    }
    \caption{Results of the user study: left (w/o \gsng) and right (w \gsng). Percentages indicate how often each method was chosen.}
    \label{tab:human_evalw}
\end{table}

\begin{figure*}[!h]
    \centering
    \begin{minipage}[t]{0.48\linewidth}
        \centering
        \input{data/illustration/quali_comparison/wo_gsn/3_4_short}
        \caption{Images generated \textit{without} \gsng (\sdun; same seeds)}
        \label{fig:wo_gsng_comparison}
    \end{minipage}
    \hfill
    \begin{minipage}[t]{0.48\linewidth}
        \centering
        \input{data/illustration/quali_comparison/sd3/3_11_short}
        \caption{Images generated with \sdtrois (same seeds for both)}
        \label{fig:sd3_qual_311}
    \end{minipage}
\end{figure*}

\begin{figure*}[!h]
    \centering
    \input{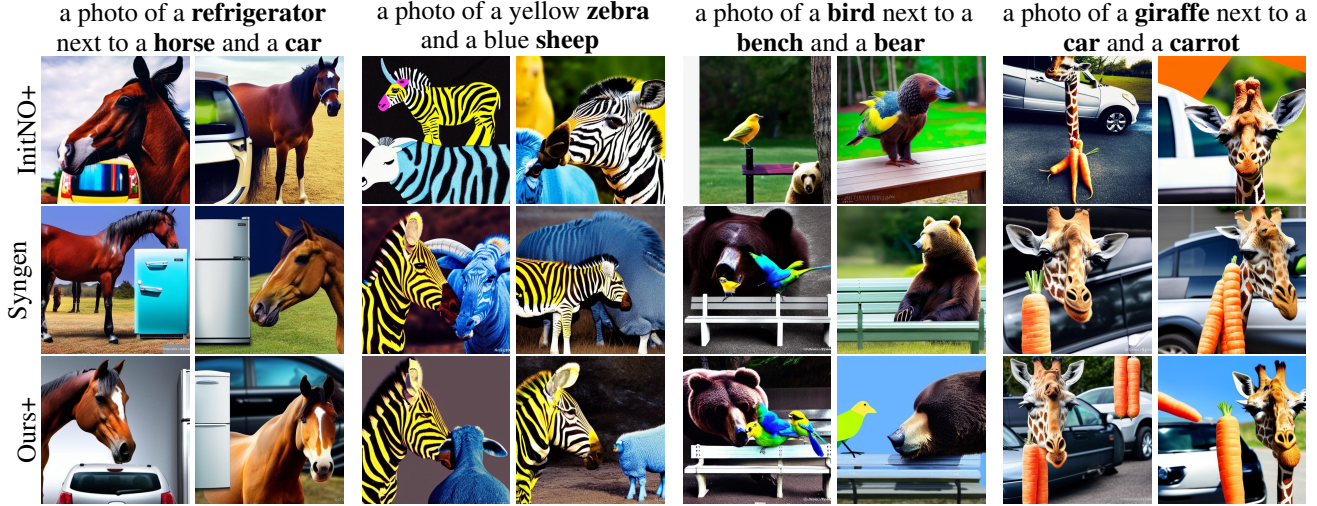}
    \caption{Images generated \textit{with} \gsng (\sdun; same seeds)}
    \label{fig:gsng_comparison}
\end{figure*}

We conducted a subjective user study to evaluate human preferences across various methods on \sdun, including 37 candidates. For each comparison, we presented images generated by each method using the same randomly selected prompt and seed, with participants asked to choose the best matches or select none if applicable. The study consisted of two phases. In the first phase, we compared InitNO with Ours, followed by a second phase where we evaluated Syngen, InitNO+, and Ours+. We present the result in \autoref{tab:human_evalw}. Our method demonstrates a significant improvement over InitNO in the one-step \iteref setup, further validating the effectiveness of our approach. Additionally, with guidance, participants chose Ours+ more frequently than the others, indicating superior alignment with the text prompts. Further details are provided in the \supmat.

\subsection{Qualitative Comparison}

We present a qualitative comparison of image generation using the same two seeds with different prompts for methods without \gsng in \autoref{fig:wo_gsng_comparison} and \autoref{fig:sd3_qual_311}. Our method better mitigates catastrophic neglect \eg InitNO struggles to clearly represent both entities in the prompt \textit{a photo of a bench and an elephant}. Even with challenging prompts containing three entities, our approach yields superior results, as the later \iteref step helps to distinguish the entities more effectively. In \autoref{fig:gsng_comparison}, we present results for methods employing \gsng. Our method significantly enhances the separation of three objects. For instance, Syngen and InitNO+ fail sometimes to generate certain entities (\eg Syngen: \textit{car} in the first prompt, \textit{bird} in the third prompt; InitNO+: \textit{refrigerator} in the first prompt, \textit{carrot} in the last prompt). Furthermore, our approach better differentiates the entities (\eg Syngen: \textit{sheep} in the second prompt are not distinguishable, while InitNO+ mixes \textit{sheep} with \textit{zebra} in the second image of the second prompt and \textit{bird} with \textit{bear} in the first image of the third prompt). Our approach demonstrates superior performance in effectively generating and distinguishing entities compared to existing approaches. We provide further examples for \sdun/3 in the \supmat.

\subsection{Study of the \iteref Placement}

\begin{figure}[!h]
    \centering
    \includegraphics[width=1\linewidth]{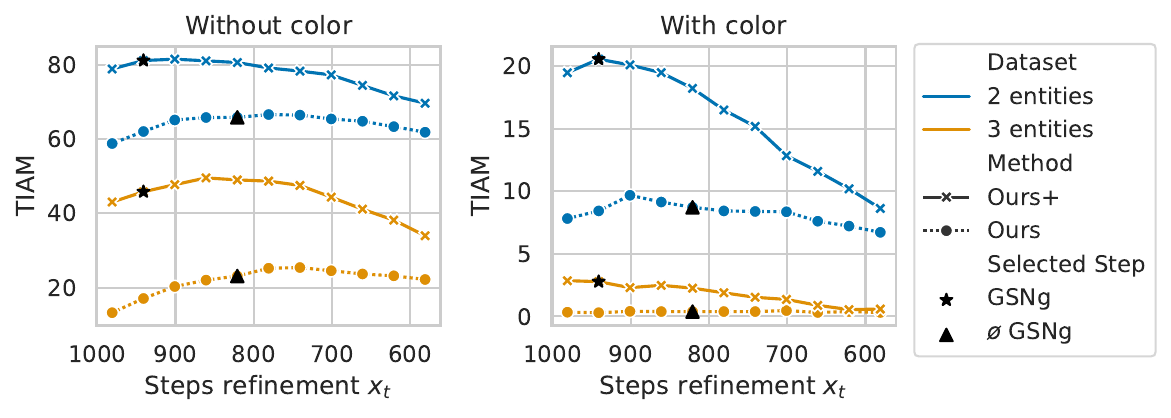}
    \caption{TIAM score according to \iteref step for entities with color (right) and w/o (left).}
    \label{fig:tiam_score_steps_objects}
\end{figure}
We conducted an exhaustive study on the optimal diffusion steps to do the \iteref step (\autoref{fig:tiam_score_steps_objects}). The candidate steps identified in \autoref{sec:exp_settings} align well with the results, as they consistently demonstrate good performance across all datasets. This reinforces the validity of our validation approach for determining \iteref step candidates: optimizing too early was less effective than making adjustments at later stages. However, delaying corrections too much is detrimental, indicating the necessity for a careful trade-off in timing when modifying the signal. We noted that the use of \gsng follows similar trends but consistently yields better results by facilitating slight, continuous adjustments to the signal. We also found that for datasets with color, one can obtain better results by setting different \iteref steps. This conclusion stems from the understanding that color modifications should be implemented early in the diffusion process, as colors appear to be defined at the outset. Making adjustments later may hinder effective integration. In contrast, modifying the signal for entities is more advantageous at later stages, allowing for greater precision in distinguishing between different entities.

\section{Limitations}

The GSN approach is constrained by the model's inherent knowledge, although we can incorporate external information through well-designed GSN losses. This limitation affects our ability to optimize effectively, as challenges persist \eg rare concept, object confusion, reasoning, counting \cite{udandarao2024zeroshot,Paiss_2023_ICCV}. Consequently, we may encounter failures due to the model's out-of-distribution behavior. Our work has demonstrated that a thorough understanding of signal construction during diffusion allows for the selection of optimization steps that enhance image generation while limiting the number of hyperparameters and the number of \iteref steps, such as optimization thresholds according to the step of diffusion. However, we believe that despite the challenges associated with testing numerous thresholds and hyperparameters, an approach utilizing well-engineered optimization thresholds could improve performance, particularly when considering signal construction. Finally, like other GSN methods, our approach requires back-propagation through the model, which is computationally intensive.

\section{Conclusion}

In this study, we improve the application of GSN criteria by exploring how the signal evolves during the denoising process. We presented a method for identifying and validating an optimal refinement step. Our findings show that while early-stage signal modifications are less effective, timely adjustments can lead to significant performance improvements, enabling the generation of semantically aligned images and achieving state-of-the-art results, as demonstrated through extensive experiments. Furthermore, this approach reduces the number of hyperparameters and \iteref compared to some state-of-the-art methods \eg InitNO, simplifying the model setup and enhancing overall efficiency. We observed that the position of the \iteref step depends on the specific elements we are seeking to correct. For example, color modifications should occur earlier in the process, while adjustments to entities can be made slightly later. Future developments of GSN methods could build on these insights by selecting refinement steps tailored to the particular aspects being adjusted. Additionally, incorporating a reminder loss \cite{Agarwal_2023_ICCV} could further enhance the approach by providing the model with a memory of the signal across sampling steps.

\section*{Acknowledgement}

This work was performed using HPC resources from GENCI-IDRIS (Grant 2022-AD011014009) and on the use of the FactoryIA supercomputer, financially supported by the Ile-de-France Regional Council. This work was partly supported by the SHARP ANR project ANR-23-PEIA-0008 in the context of the France 2030 program.

\newpage
\bibliography{ref}

\begin{thebibliography}{}

\bibitem[\protect\citeauthoryear{Agarwal \bgroup \em et al.\egroup
  }{2023}]{Agarwal_2023_ICCV}
Aishwarya Agarwal, Srikrishna Karanam, K~J Joseph, Apoorv Saxena, Koustava
  Goswami, and Balaji~Vasan Srinivasan.
\newblock A-star: Test-time attention segregation and retention for
  text-to-image synthesis.
\newblock In {\em Proceedings of the IEEE/CVF International Conference on
  Computer Vision (ICCV)}, pages 2283--2293, October 2023.

\bibitem[\protect\citeauthoryear{Balaji \bgroup \em et al.\egroup
  }{2023}]{balaji2023ediffi}
Yogesh Balaji, Seungjun Nah, Xun Huang, Arash Vahdat, Jiaming Song, Qinsheng
  Zhang, Karsten Kreis, Miika Aittala, Timo Aila, Samuli Laine, Bryan
  Catanzaro, Tero Karras, and Ming-Yu Liu.
\newblock {eDiff-I: Text-to-Image Diffusion Models with an Ensemble of Expert
  Denoisers}.
\newblock {\em arXiv 2211.01324}, 2023.

\bibitem[\protect\citeauthoryear{Chefer \bgroup \em et al.\egroup
  }{2023}]{chefer2023attendandexcite}
Hila Chefer, Yuval Alaluf, Yael Vinker, Lior Wolf, and Daniel Cohen-Or.
\newblock Attend-and-excite: Attention-based semantic guidance for
  text-to-image diffusion models.
\newblock {\em ACM Trans. Graph.}, 42(4), jul 2023.

\bibitem[\protect\citeauthoryear{Chen \bgroup \em et al.\egroup
  }{2023}]{chen2023pixartalpha}
Junsong Chen, Jincheng Yu, Chongjian Ge, Lewei Yao, Enze Xie, Yue Wu, Zhongdao
  Wang, James Kwok, Ping Luo, Huchuan Lu, and Zhenguo Li.
\newblock Pixart-$\alpha$: Fast training of diffusion transformer for
  photorealistic text-to-image synthesis, 2023.

\bibitem[\protect\citeauthoryear{Chen \bgroup \em et al.\egroup
  }{2024a}]{chen2024pixartsigma}
Junsong Chen, Chongjian Ge, Enze Xie, Yue Wu, Lewei Yao, Xiaozhe Ren, Zhongdao
  Wang, Ping Luo, Huchuan Lu, and Zhenguo Li.
\newblock Pixart-$\sigma$: Weak-to-strong training of diffusion transformer for
  4k text-to-image generation, 2024.

\bibitem[\protect\citeauthoryear{Chen \bgroup \em et al.\egroup
  }{2024b}]{Chen_2024_WACV}
Minghao Chen, Iro Laina, and Andrea Vedaldi.
\newblock Training-free layout control with cross-attention guidance.
\newblock In {\em Proceedings of the IEEE/CVF Winter Conference on Applications
  of Computer Vision (WACV)}, pages 5343--5353, January 2024.

\bibitem[\protect\citeauthoryear{Chen}{2023}]{chen2023importance}
Ting Chen.
\newblock On the importance of noise scheduling for diffusion models, 2023.

\bibitem[\protect\citeauthoryear{Choi \bgroup \em et al.\egroup
  }{2021}]{Choi_2021_ICCV}
Jooyoung Choi, Sungwon Kim, Yonghyun Jeong, Youngjune Gwon, and Sungroh Yoon.
\newblock Ilvr: Conditioning method for denoising diffusion probabilistic
  models.
\newblock In {\em Proceedings of the IEEE/CVF International Conference on
  Computer Vision (ICCV)}, pages 14367--14376, October 2021.

\bibitem[\protect\citeauthoryear{Choi \bgroup \em et al.\egroup
  }{2022}]{Choi_2022_CVPR}
Jooyoung Choi, Jungbeom Lee, Chaehun Shin, Sungwon Kim, Hyunwoo Kim, and
  Sungroh Yoon.
\newblock Perception prioritized training of diffusion models.
\newblock In {\em Proceedings of the IEEE/CVF Conference on Computer Vision and
  Pattern Recognition (CVPR)}, pages 11472--11481, June 2022.

\bibitem[\protect\citeauthoryear{Esser \bgroup \em et al.\egroup
  }{2024}]{esser2024scalingrectifiedflowtransformers}
Patrick Esser, Sumith Kulal, Andreas Blattmann, Rahim Entezari, Jonas Müller,
  Harry Saini, Yam Levi, Dominik Lorenz, Axel Sauer, Frederic Boesel, Dustin
  Podell, Tim Dockhorn, Zion English, Kyle Lacey, Alex Goodwin, Yannik Marek,
  and Robin Rombach.
\newblock Scaling rectified flow transformers for high-resolution image
  synthesis, 2024.

\bibitem[\protect\citeauthoryear{Everaert \bgroup \em et al.\egroup
  }{2024}]{Everaert_2024_WACV}
Martin~Nicolas Everaert, Athanasios Fitsios, Marco Bocchio, Sami Arpa, Sabine
  Süsstrunk, and Radhakrishna Achanta.
\newblock {E}xploiting the {S}ignal-{L}eak {B}ias in {D}iffusion {M}odels.
\newblock In {\em Proceedings of the IEEE/CVF Winter Conference on Applications
  of Computer Vision (WACV)}, pages 4025--4034, January 2024.

\bibitem[\protect\citeauthoryear{Feng \bgroup \em et al.\egroup
  }{2023}]{feng2023trainingfreestructureddiffusionguidance}
Weixi Feng, Xuehai He, Tsu-Jui Fu, Varun Jampani, Arjun Akula, Pradyumna
  Narayana, Sugato Basu, Xin~Eric Wang, and William~Yang Wang.
\newblock Training-free structured diffusion guidance for compositional
  text-to-image synthesis, 2023.

\bibitem[\protect\citeauthoryear{Fleiss and others}{1971}]{fleiss1971mns}
J.L. Fleiss et~al.
\newblock {Measuring nominal scale agreement among many raters}.
\newblock {\em Psychological Bulletin}, 76(5):378--382, 1971.

\bibitem[\protect\citeauthoryear{Grimal \bgroup \em et al.\egroup
  }{2024}]{Grimal_2024_WACV}
Paul Grimal, Herv\'e Le~Borgne, Olivier Ferret, and Julien Tourille.
\newblock Tiam - a metric for evaluating alignment in text-to-image generation.
\newblock In {\em Proceedings of the IEEE/CVF Winter Conference on Applications
  of Computer Vision (WACV)}, pages 2890--2899, January 2024.

\bibitem[\protect\citeauthoryear{Grimal \bgroup \em et al.\egroup
  }{2025}]{grimal2025signal}
Paul Grimal, Herv{\'e}~Le Borgne, and Olivier Ferret.
\newblock Signal dynamics in diffusion models: Enhancing text-to-image
  alignment through step selection, 2025.

\bibitem[\protect\citeauthoryear{Guo \bgroup \em et al.\egroup
  }{2024}]{guo2024initno}
Xiefan Guo, Jinlin Liu, Miaomiao Cui, Jiankai Li, Hongyu Yang, and Di~Huang.
\newblock Initno: Boosting text-to-image diffusion models via initial noise
  optimization, 2024.

\bibitem[\protect\citeauthoryear{Hertz \bgroup \em et al.\egroup
  }{2022}]{hertz2022prompttopromptimageeditingcross}
Amir Hertz, Ron Mokady, Jay Tenenbaum, Kfir Aberman, Yael Pritch, and Daniel
  Cohen-Or.
\newblock Prompt-to-prompt image editing with cross attention control, 2022.

\bibitem[\protect\citeauthoryear{Ho and Salimans}{2022}]{ho2022classifierfree}
Jonathan Ho and Tim Salimans.
\newblock Classifier-free diffusion guidance.
\newblock {\em arXiv 2207.12598}, 2022.

\bibitem[\protect\citeauthoryear{Ho \bgroup \em et al.\egroup
  }{2020}]{ho2020denoising}
Jonathan Ho, Ajay Jain, and Pieter Abbeel.
\newblock Denoising diffusion probabilistic models.
\newblock {\em arXiv preprint arxiv:2006.11239}, 2020.

\bibitem[\protect\citeauthoryear{Jocher \bgroup \em et al.\egroup
  }{2023}]{yolov8_ultralytics}
Glenn Jocher, Ayush Chaurasia, and Jing Qiu.
\newblock Yolo by ultralytics, 2023.

\bibitem[\protect\citeauthoryear{Kingma and
  Ba}{2017}]{kingma2017adammethodstochasticoptimization}
Diederik~P. Kingma and Jimmy Ba.
\newblock Adam: A method for stochastic optimization, 2017.

\bibitem[\protect\citeauthoryear{Kwon \bgroup \em et al.\egroup
  }{2023}]{kwon2023diffusionmodelssemanticlatent}
Mingi Kwon, Jaeseok Jeong, and Youngjung Uh.
\newblock Diffusion models already have a semantic latent space, 2023.

\bibitem[\protect\citeauthoryear{Landis and Koch}{1977}]{Landis77}
J.~Richard Landis and Gary~G. Koch.
\newblock The measurement of observer agreement for categorical data.
\newblock {\em Biometrics}, 33(1), 1977.

\bibitem[\protect\citeauthoryear{Li \bgroup \em et al.\egroup
  }{2022}]{li2022blip}
Junnan Li, Dongxu Li, Caiming Xiong, and Steven Hoi.
\newblock Blip: Bootstrapping language-image pre-training for unified
  vision-language understanding and generation.
\newblock In {\em ICML}, 2022.

\bibitem[\protect\citeauthoryear{Li \bgroup \em et al.\egroup
  }{2023a}]{li2023gligen}
Yuheng Li, Haotian Liu, Qingyang Wu, Fangzhou Mu, Jianwei Yang, Jianfeng Gao,
  Chunyuan Li, and Yong~Jae Lee.
\newblock Gligen: Open-set grounded text-to-image generation.
\newblock {\em CVPR}, 2023.

\bibitem[\protect\citeauthoryear{Li \bgroup \em et al.\egroup
  }{2023b}]{li2023divide}
Yumeng Li, Margret Keuper, Dan Zhang, and Anna Khoreva.
\newblock Divide \& bind your attention for improved generative semantic
  nursing.
\newblock In {\em 34th British Machine Vision Conference 2023, {BMVC} 2023},
  2023.

\bibitem[\protect\citeauthoryear{Lin \bgroup \em et al.\egroup
  }{2014}]{lin2015microsoft}
Tsung-Yi Lin, Michael Maire, Serge Belongie, James Hays, Pietro Perona, Deva
  Ramanan, Piotr Doll{\'a}r, and C.~Lawrence Zitnick.
\newblock Microsoft coco: Common objects in context.
\newblock In David Fleet, Tomas Pajdla, Bernt Schiele, and Tinne Tuytelaars,
  editors, {\em Computer Vision -- ECCV 2014}, pages 740--755, Cham, 2014.
  Springer International Publishing.

\bibitem[\protect\citeauthoryear{Lin \bgroup \em et al.\egroup
  }{2024}]{Lin_2024_WACV}
Shanchuan Lin, Bingchen Liu, Jiashi Li, and Xiao Yang.
\newblock Common diffusion noise schedules and sample steps are flawed.
\newblock In {\em Proceedings of the IEEE/CVF Winter Conference on Applications
  of Computer Vision (WACV)}, pages 5404--5411, January 2024.

\bibitem[\protect\citeauthoryear{Lipman \bgroup \em et al.\egroup
  }{2023}]{lipman2023flow}
Yaron Lipman, Ricky T.~Q. Chen, Heli Ben-Hamu, Maximilian Nickel, and Matthew
  Le.
\newblock Flow matching for generative modeling.
\newblock In {\em The Eleventh International Conference on Learning
  Representations}, 2023.

\bibitem[\protect\citeauthoryear{Meng \bgroup \em et al.\egroup
  }{2022}]{meng2022sdedit}
Chenlin Meng, Yutong He, Yang Song, Jiaming Song, Jiajun Wu, Jun-Yan Zhu, and
  Stefano Ermon.
\newblock {SDE}dit: Guided image synthesis and editing with stochastic
  differential equations.
\newblock In {\em International Conference on Learning Representations}, 2022.

\bibitem[\protect\citeauthoryear{Mou \bgroup \em et al.\egroup
  }{2023}]{mou2023t2i}
Chong Mou, Xintao Wang, Liangbin Xie, Yanze Wu, Jian Zhang, Zhongang Qi, Ying
  Shan, and Xiaohu Qie.
\newblock T2i-adapter: Learning adapters to dig out more controllable ability
  for text-to-image diffusion models.
\newblock {\em arXiv preprint arXiv:2302.08453}, 2023.

\bibitem[\protect\citeauthoryear{Paiss \bgroup \em et al.\egroup
  }{2023}]{Paiss_2023_ICCV}
Roni Paiss, Ariel Ephrat, Omer Tov, Shiran Zada, Inbar Mosseri, Michal Irani,
  and Tali Dekel.
\newblock Teaching clip to count to ten.
\newblock In {\em Proceedings of the IEEE/CVF International Conference on
  Computer Vision (ICCV)}, pages 3170--3180, October 2023.

\bibitem[\protect\citeauthoryear{Park \bgroup \em et al.\egroup
  }{2023}]{park2023understanding}
Yong-Hyun Park, Mingi Kwon, Jaewoong Choi, Junghyo Jo, and Youngjung Uh.
\newblock Understanding the latent space of diffusion models through the lens
  of riemannian geometry.
\newblock In {\em Thirty-seventh Conference on Neural Information Processing
  Systems}, 2023.

\bibitem[\protect\citeauthoryear{Peebles and Xie}{2022}]{Peebles2022DiT}
William Peebles and Saining Xie.
\newblock Scalable diffusion models with transformers.
\newblock {\em arXiv preprint arXiv:2212.09748}, 2022.

\bibitem[\protect\citeauthoryear{Podell \bgroup \em et al.\egroup
  }{2023}]{podell2023sdxl}
Dustin Podell, Zion English, Kyle Lacey, Andreas Blattmann, Tim Dockhorn, Jonas
  Müller, Joe Penna, and Robin Rombach.
\newblock Sdxl: Improving latent diffusion models for high-resolution image
  synthesis, 2023.

\bibitem[\protect\citeauthoryear{Radford \bgroup \em et al.\egroup
  }{2021}]{radford2021learning}
Alec Radford, Jong~Wook Kim, Chris Hallacy, Aditya Ramesh, Gabriel Goh,
  Sandhini Agarwal, Girish Sastry, Amanda Askell, Pamela Mishkin, Jack Clark,
  Gretchen Krueger, and Ilya Sutskever.
\newblock Learning transferable visual models from natural language
  supervision, 2021.

\bibitem[\protect\citeauthoryear{Raffel \bgroup \em et al.\egroup
  }{2020}]{raffel2020exploring}
Colin Raffel, Noam Shazeer, Adam Roberts, Katherine Lee, Sharan Narang, Michael
  Matena, Yanqi Zhou, Wei Li, and Peter~J. Liu.
\newblock Exploring the limits of transfer learning with a unified text-to-text
  transformer.
\newblock {\em arXiv 1910.10683}, 2020.

\bibitem[\protect\citeauthoryear{Ramesh \bgroup \em et al.\egroup
  }{2022}]{ramesh2022hierarchical}
Aditya Ramesh, Prafulla Dhariwal, Alex Nichol, Casey Chu, and Mark Chen.
\newblock Hierarchical text-conditional image generation with clip latents.
\newblock {\em arXiv 2204.06125}, 2022.

\bibitem[\protect\citeauthoryear{Rassin \bgroup \em et al.\egroup
  }{2023}]{rassin2023linguistic}
Royi Rassin, Eran Hirsch, Daniel Glickman, Shauli Ravfogel, Yoav Goldberg, and
  Gal Chechik.
\newblock Linguistic binding in diffusion models: Enhancing attribute
  correspondence through attention map alignment.
\newblock In {\em Thirty-seventh Conference on Neural Information Processing
  Systems}, 2023.

\bibitem[\protect\citeauthoryear{Rissanen \bgroup \em et al.\egroup
  }{2023}]{rissanen2023generative}
Severi Rissanen, Markus Heinonen, and Arno Solin.
\newblock Generative modelling with inverse heat dissipation.
\newblock In {\em International Conference on Learning Representations (ICLR)},
  2023.

\bibitem[\protect\citeauthoryear{Rombach \bgroup \em et al.\egroup
  }{2022}]{rombach2021highresolution}
Robin Rombach, Andreas Blattmann, Dominik Lorenz, Patrick Esser, and Bj\"orn
  Ommer.
\newblock High-resolution image synthesis with latent diffusion models.
\newblock In {\em Proceedings of the IEEE/CVF Conference on Computer Vision and
  Pattern Recognition (CVPR)}, pages 10684--10695, June 2022.

\bibitem[\protect\citeauthoryear{Saharia \bgroup \em et al.\egroup
  }{2022}]{saharia2022photorealistic}
Chitwan Saharia, William Chan, Saurabh Saxena, Lala Li, Jay Whang, Emily
  Denton, Seyed Kamyar~Seyed Ghasemipour, Raphael Gontijo-Lopes, Burcu~Karagol
  Ayan, Tim Salimans, Jonathan Ho, David~J. Fleet, and Mohammad Norouzi.
\newblock Photorealistic text-to-image diffusion models with deep language
  understanding.
\newblock In Alice~H. Oh, Alekh Agarwal, Danielle Belgrave, and Kyunghyun Cho,
  editors, {\em Advances in Neural Information Processing Systems}, 2022.

\bibitem[\protect\citeauthoryear{Schuhmann \bgroup \em et al.\egroup
  }{2022}]{schuhmann2022laion5bopenlargescaledataset}
Christoph Schuhmann, Romain Beaumont, Richard Vencu, Cade Gordon, Ross
  Wightman, Mehdi Cherti, Theo Coombes, Aarush Katta, Clayton Mullis, Mitchell
  Wortsman, Patrick Schramowski, Srivatsa Kundurthy, Katherine Crowson, Ludwig
  Schmidt, Robert Kaczmarczyk, and Jenia Jitsev.
\newblock Laion-5b: An open large-scale dataset for training next generation
  image-text models, 2022.

\bibitem[\protect\citeauthoryear{Segalis \bgroup \em et al.\egroup
  }{2023}]{segalis2023picture}
Eyal Segalis, Dani Valevski, Danny Lumen, Yossi Matias, and Yaniv Leviathan.
\newblock A picture is worth a thousand words: Principled recaptioning improves
  image generation, 2023.

\bibitem[\protect\citeauthoryear{Sun \bgroup \em et al.\egroup
  }{2024}]{sun2024spatialaware}
Wenqiang Sun, Teng Li, Zehong Lin, and Jun Zhang.
\newblock Spatial-aware latent initialization for controllable image
  generation, 2024.

\bibitem[\protect\citeauthoryear{Tang \bgroup \em et al.\egroup
  }{2023}]{tang2023daam}
Raphael Tang, Linqing Liu, Akshat Pandey, Zhiying Jiang, Gefei Yang, Karun
  Kumar, Pontus Stenetorp, Jimmy Lin, and Ferhan Ture.
\newblock What the {DAAM}: Interpreting stable diffusion using cross attention.
\newblock In {\em Proceedings of the 61st Annual Meeting of the Association for
  Computational Linguistics (Volume 1: Long Papers)}, 2023.

\bibitem[\protect\citeauthoryear{Udandarao \bgroup \em et al.\egroup
  }{2024}]{udandarao2024zeroshot}
Vishaal Udandarao, Ameya Prabhu, Adhiraj Ghosh, Yash Sharma, Philip H.~S. Torr,
  Adel Bibi, Samuel Albanie, and Matthias Bethge.
\newblock No "zero-shot" without exponential data: Pretraining concept
  frequency determines multimodal model performance, 2024.

\bibitem[\protect\citeauthoryear{Xie \bgroup \em et al.\egroup
  }{2023}]{Xie_2023_ICCV}
Jinheng Xie, Yuexiang Li, Yawen Huang, Haozhe Liu, Wentian Zhang, Yefeng Zheng,
  and Mike~Zheng Shou.
\newblock Boxdiff: Text-to-image synthesis with training-free box-constrained
  diffusion.
\newblock In {\em Proceedings of the IEEE/CVF International Conference on
  Computer Vision (ICCV)}, pages 7452--7461, 2023.

\bibitem[\protect\citeauthoryear{Yu \bgroup \em et al.\egroup
  }{2023}]{yu2023freedom}
Jiwen Yu, Yinhuai Wang, Chen Zhao, Bernard Ghanem, and Jian Zhang.
\newblock Freedom: Training-free energy-guided conditional diffusion model.
\newblock {\em Proceedings of the IEEE/CVF International Conference on Computer
  Vision (ICCV)}, 2023.

\bibitem[\protect\citeauthoryear{Yuksekgonul \bgroup \em et al.\egroup
  }{2023}]{yuksekgonul2023when}
Mert Yuksekgonul, Federico Bianchi, Pratyusha Kalluri, Dan Jurafsky, and James
  Zou.
\newblock When and why vision-language models behave like bags-of-words, and
  what to do about it?
\newblock In {\em The Eleventh International Conference on Learning
  Representations}, 2023.

\bibitem[\protect\citeauthoryear{Zhang \bgroup \em et al.\egroup
  }{2023}]{zhang2023adding}
Lvmin Zhang, Anyi Rao, and Maneesh Agrawala.
\newblock Adding conditional control to text-to-image diffusion models, 2023.

\end{thebibliography}
\bibliographystyle{data/template/ijcai2025/named}

\clearpage
\appendix
\section{Appendix}

The appendix is summarized as follows:

\begin{itemize}
    \item Section~\ref{sec:implementation}: detailed descriptions of the implementations and methods used,
    \item Section~\ref{sec:sd3_details}: description of the adaptation of GSN to Stable Diffusion 3 and experimental results results on validation datasets,
    \item Section~\ref{sec:tiam}: an overview of the TIAM evaluation process,
    \item Section~\ref{sec:eval_attend}: a summary of the evaluation framework from Attend\&Excite, along with additional results,
    \item Section~\ref{sec:user_study}: detailed information about the user study,
    \item Section~\ref{sec:qual_samples}: additional comparative sample outputs,
    \item Section~\ref{sec:results}: values for the figures in the main document and supplementary results, including Section \ref{sec:validation} for the validation set and Section \ref{sec:test} for the test set.
\end{itemize}

\clearpage
\subsection{Text-to-Image Methods Setup For Stable Diffusion 1.4}
\label{sec:implementation}
We provide here some implementation details about Stable Diffusion 1.4. and methods used.

\paragraph{Stable-Diffusion version 1.4 (SD 1.4)} We use the model hosted on HuggingFace\footnote{\url{https://huggingface.co/CompVis/stable-diffusion-v1-4}} with the DDPM Scheduler\footnote{\url{https://huggingface.co/docs/diffusers/api/schedulers/ddpm}} and 50 sampling steps. All the methods were performed with a Classifier Free Guidance \citep{ho2022classifierfree} of 7.5.

\paragraph{Attend\&Excite}

We utilize the implementation provided by the Diffusers library\footnote{\url{https://huggingface.co/docs/diffusers/api/pipelines/attend_and_excite}}. The iterative refinement occurs at the sampling steps 0, 10, and 20, where the loss must reach specified thresholds of 0.05, 0.5, and 0.8, respectively. A maximum of 20 iterative refinement steps is performed. The learning rate decreases progressively with each sampling step, starting at an initial value of 20. They perform the GSN guidance for the 25 first sampling steps.

\paragraph{Divide and Bind}

We utilize the official implementation\footnote{\url{https://github.com/boschresearch/Divide-and-Bind}}. We follow the authors' recommendation and use the \textit{tv} loss for the prompts without colors and the \textit{tv bind} loss for the prompts with colors. The iterative refinement occurs at the sampling steps 0, 10, and 20, where the losses must reach specified thresholds of 0.05, 0.2, and 0.3, respectively. A maximum of 50 iterative refinement steps is performed. The learning rate decreases progressively with each sampling step, starting at an initial value of 20. They perform the GSN guidance for the 25 first sampling steps.

\paragraph{InitNO}

We utilize the official implementation provided in the repository\footnote{\url{https://github.com/xiefan-guo/initno}}. The authors designed a loss function comprising three components: self-attention loss, cross-attention loss, and KL divergence loss.
During the multi-round step, an iterative refinement is performed. If the defined thresholds for the cross-attention and self-attention losses are not met, the optimization is repeated by sampling a new starting latent, up to a maximum of five attempts. If the objectives remain unattainable, inference is conducted using the optimized starting latent representation that achieves the best score relative to the objectives. The KL divergence loss is applied exclusively during the boosting step, where optimization is performed after each back-propagation on the attention losses to ensure that the starting latent image remains within an appropriate interval.
Iterative refinement steps are also conducted at sampling steps 10 and 20. For both the boosting step and iterative refinement, the losses must meet specified thresholds of 0.2 for the cross-attention loss and 0.3 for the self-attention loss. The learning rate decreases progressively with each sampling step, beginning at an initial value of 20. Additionally, GSN guidance is applied for the first 25 sampling steps.

Additionally, we discovered in the code that the implementation includes a \textit{clean cross-attention loss}, which applies a specialized processing of the attention maps using Otsu thresholding during the multi-round step and \gsng. The code also incorporates a \textit{cross-attention alignment loss} for the \gsng, seemingly designed to encourage consistency in token activation zones across diffusion steps. To the best of our knowledge, these details are not mentioned in the main paper.

\paragraph{Syngen}
We utilize the official implementation\footnote{\url{https://github.com/RoyiRa/Linguistic-Binding-in-Diffusion-Models}}. They apply only a GSN guidance for the first 25 sampling steps. They use a learning rate of 20.

Syngen is designed to accept prompts that consist solely of entities with attributes. For instance, when the prompt is ``a photo of a cat and a dog'', the cross-attention maps corresponding to ``a photo of'' are utilized. To enhance the results, we remove the cross-attention maps associated with the initial tokens. This adjustment led to an approximate increase of 1 in performance during the experiments. The scores reported for Syngen in the paper reflect these beneficial modifications.

\begin{table*}
    \centering
    \begin{tabular}{cccccc}
    \toprule
    \multirow[c]{2}{*}{\rotatebox{0}{\shortstack{$n$ shift of                                               \\latent image}}} &\multirow[c]{2}{*}{\rotatebox{0}{Methods}} & \multicolumn{2}{c}{w/o colors} & \multicolumn{2}{c}{with colors} \\
       &         & 2 entities           & 3 entities           & 2 entities           & 3 entities          \\
    \midrule
    20 & Syngen+ & $77.81_{33.98/5.38}$ & $36.17_{36.32/5.39}$ & $20.08_{37.07/5.3}$  & $1.88_{36.85/5.27}$ \\
    \midrule
    50 & Syngen+ & $75.81_{33.78/5.33}$ & $36.23_{36.17/5.35}$ & $18.23_{37.06/5.26}$ & $1.9_{36.97/5.25}$  \\
    \bottomrule
\end{tabular}
    \caption{TIAM score on the different datasets with Syngen and an iterative refinement step using the Syngen criterion.}
    \label{tab:syngen}
\end{table*}

We attempted to introduce one refinement step for Syngen. Specifically, we applied a refinement step at the first sampling step, similar to InitNO, and conducted 20 and 50 optimization iterations using the loss function of Syngen. The Adam optimizer was employed with a learning rate of $1\times10^{-2}$. The TIAM scores are reported in \autoref{tab:syngen}. However, we did not achieve better results compared to configurations without refinement steps. While improvements may be possible, further research is required to identify optimal hyperparameters.

\paragraph{Ours}
Following the Attend\&Excite framework, we apply Gaussian smoothing to the attention maps using a kernel size of 3 and a standard deviation of 0.5. During the iterative refinement step, we conduct 50 latent image shifts without aiming to achieve a specific threshold. For the configuration utilizing GSN guidance, we incorporate the Syngen GSN guidance after proceeding with the iterative refinement step.

\clearpage
\subsection{Stable Diffusion 3}
\label{sec:sd3_details}
We use the implementation available on Hugging Face\footnote{\url{https://huggingface.co/stabilityai/stable-diffusion-3-medium-diffusers}} with the default scheduler, FlowMatchEulerDiscreteScheduler\footnote{\url{https://huggingface.co/docs/diffusers/api/schedulers/flow_match_euler_discrete}}, configured with 28 sampling steps, a Classifier-Free Guidance \citep{ho2022classifierfree} of 7.0, and bfloat16 precision for image generation.  For \iteref, we apply an Adam optimizer with a learning rate of $1\times10^{-2}$ and 50 steps of optimization.

Stable Diffusion 3 (SD 3) is a Flow Matching model designed to construct a probabilistic path between two distributions, $p_0$ and $p_1$, where $p_0$ is the target distribution and  $p_1 \sim \mathcal{N}(0, I)$. The model learns to transport points from one distribution to another. The latent image transport path can be interpreted as a denoising process, with noise progressively removed in a manner analogous to image destruction. Specifically, the latent image $x_t$ is sampled using the reparameterization trick, involving the interpolation of the image and noise. As demonstrated by Rissanen~\citet{rissanen2023generative}, isotropic noise suppresses frequency components in the data that have a lower power spectral density than the variance of the noise. Consequently, the model initially reconstructs lower frequencies and subsequently refines higher frequencies, similar to the process observed in diffusion models. During denoising, the signal can be refined to ensure alignment with the desired output using the GSN approach. Additionally, our method can be applied to select the optimal step in the denoising process. While feature extraction in Stable Diffusion models 1.4 and 1.5 is well-documented, to the best of our knowledge, this has not been extensively explored for Stable Diffusion 3, which uses a transformer-based architecture. In this architecture, T5 and CLIP serve as two distinct encoders for guidance. The model incorporates two independent transformers, each operating within its own modality space (image patches and text), while taking the other modality into account when processing the attention. We first describe how we process and extract attention maps and secondly, how we select a potential nice step to refine the latent image.

\paragraph{Extraction of the attention maps}

Stable Diffusion 3 consists of 24 transformer blocks. The latent image, represented as $x_t \in \mathbb{R}^{H \times W \times c}$, where $c$ is the number of channels in the latent space, and $H, W$ are the height and width, is patchified to produce a sequence of tokens $z \in \mathbb{R}^{hw \times d}$, where $hw = \frac{1}{2} H \times \frac{1}{2} W$, and $d$ is the token embedding dimension.

The textual embedding $t$ is formed by concatenating the embeddings from CLIP and T5 and projecting them into the same dimension $d$. This results in $t \in \mathbb{R}^{(n_{\text{CLIP}} + n_{\text{T5}}) \times d}$, where $n_{\text{CLIP}}$ and $n_{\text{T5}}$ represent the number of tokens from CLIP and T5, respectively.

When processing attention, the resulting attention maps $A$ are of size $A \in \mathbb{R}^{(hw + n_{\text{CLIP}} + n_{\text{T5}})^2 \times n_{\text{head}}}$, where $n_{\text{head}}$ is the number of attention heads. We extract the attention maps and focus on the subset where the image patches serve as the queries, and the text embeddings act as the keys. This subset is crucial as it captures the relationship between the image latent and textual concepts, ensuring the signal within the latent image aligns with the semantic meaning of the tokens.

\begin{figure*}
    \centering

    \includegraphics[height = 0.4\linewidth]{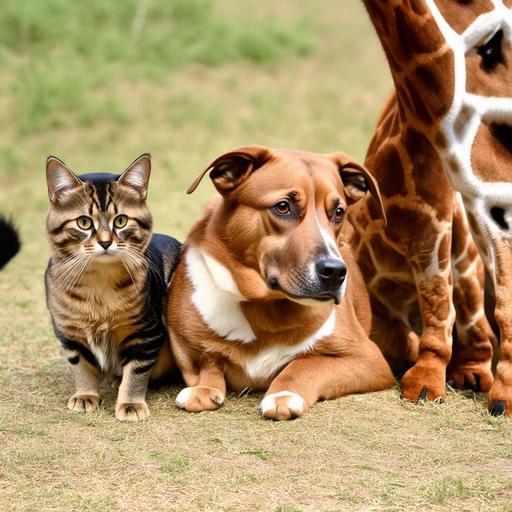}  \includegraphics[height = 0.4 \linewidth]{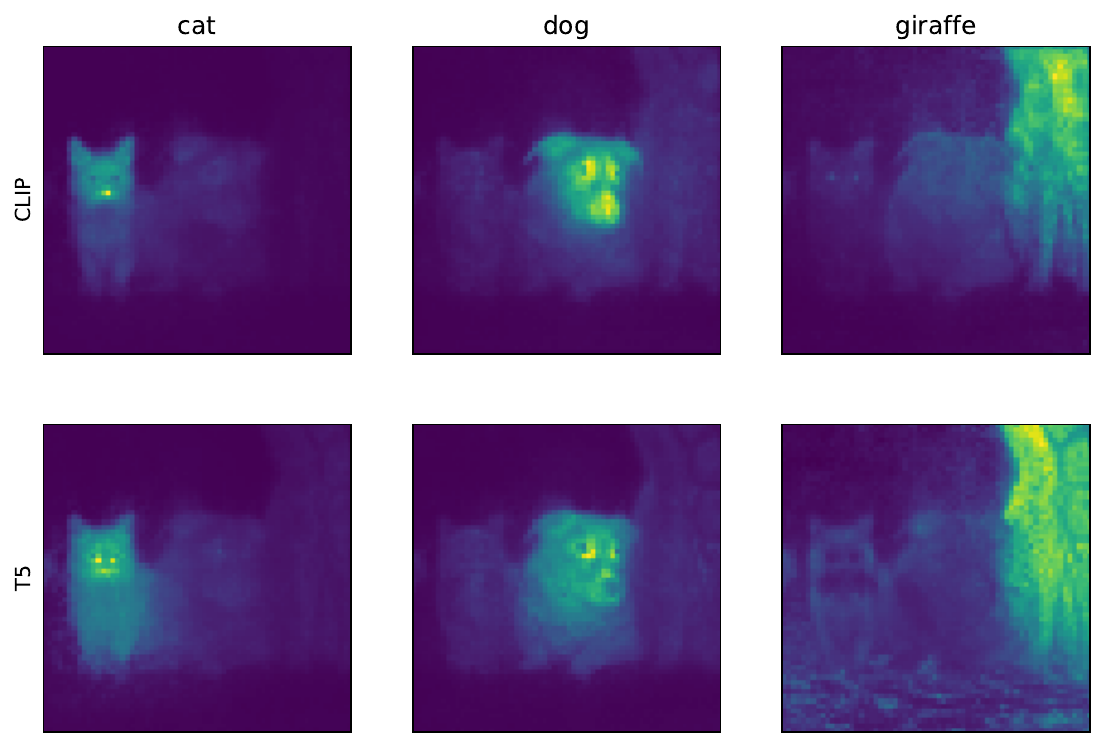}
    \caption{On the left, the image generated by Stable Diffusion 3 for the prompt ``a photo of a cat next to a dog and a giraffe''. On the right, extracted attention maps for CLIP and T5 tokens, averaged across all diffusion steps and transformer blocks. For words represented by multiple tokens, the attention maps are further averaged.}
    \label{fig:sd3_attention_maps}
\end{figure*}

To simplify the attention maps, we average across the attention heads and transformer blocks, yielding $A \in \mathbb{R}^{hw \times (n_{\text{CLIP}} + n_{\text{T5}})}$. We further refine these maps by excluding the special tokens (e.g., the start and end tokens) for both CLIP and T5, as these tend to dominate the attention distribution without contributing meaningful semantic information. The attention maps are reweighted using a softmax operation and Gaussian smoothing, as proposed by Chefer~\citet{chefer2023attendandexcite} for Stable Diffusion 1.4. For subject tokens that span multiple tokens (e.g., due to subword tokenization), we average their respective attention maps. Finally, the attention maps from CLIP and T5 are aligned and combined by averaging, producing the final attention maps, $A \in \mathbb{R}^{hw \times \mathcal{S}}$, used to guide the latent space adjustment. The loss function described in the main paper is applied to modify the latent representation accordingly. However, further investigation is required to determine whether extracting attention maps from all transformer blocks is necessary. Preliminary observations suggest that the first and last transformer blocks lack clear semantic correspondence with spatial features, as revealed through visualizations. A selective approach to choosing transformer blocks, based on a detailed analysis, could lead to more effective results. This refined attention extraction process may also serve as a foundation for future work in developing semantic map extraction techniques~\cite{tang2023daam}. We include an example of the extracted attention maps without any processing, averaged across blocks and generation steps in the \autoref{fig:sd3_attention_maps}. The semantic correspondence between text representations and visual representations can be observed.

\paragraph{Optimal \iteref step selection}

\begin{figure}
    \centering
    \includegraphics[width = 0.9\linewidth]{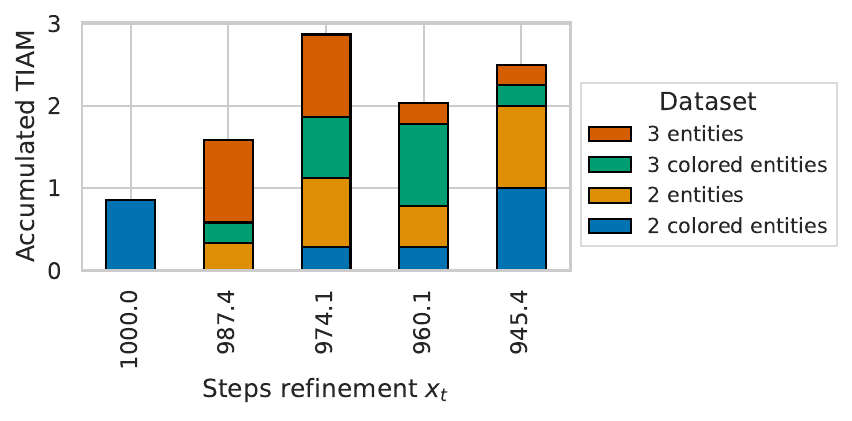}
    \caption{Accumulated TIAM scores for Stable Diffusion 3.}
    \label{fig:val_sd3}
\end{figure}

We evaluate the first 5 sampling steps. For each validation dataset, we generate 8 images per prompt using the same 8 seeds and compute the TIAM score. The scores are standardized using a min-max scaler for each dataset, and the \textit{accumulated standardized TIAM} across the \iteref steps is shown in \autoref{fig:val_sd3}. The third sampling step is optimal as it performs consistently well across all datasets. The non-standardized values are presented in \autoref{fig:val_sd3}.

\begin{table}
    \centering
    \begin{tabular}{ccccc}
\toprule
\multirow[c]{2}{*}{\rotatebox{0}{\shortstack{step of iterative\\refinement}}} & \multicolumn{2}{c}{2 entities} & \multicolumn{2}{c}{3 entities} \\
 & wo color & color & wo color & color\\
\midrule
945.4 & 88.75 & 28.75 & 62.50 & 13.75 \\
960.1 & 85.00 & 22.50 & 62.50 & 17.50 \\
974.1 & 87.50 & 22.50 & 66.25 & 16.25 \\
987.4 & 83.75 & 20.00 & 66.25 & 13.75 \\
1000 & 81.25 & 27.50 & 61.25 & 12.50 \\
\bottomrule
\end{tabular}
    \label{tab:val_value_sd3}
    \caption{TIAM score for Stable Diffusion 3 according to the steps used for the refinement with datasets with 10 prompts (when color TIAM score ground truth colors that is displayed).}
\end{table}

\clearpage
\subsection{TIAM}
\label{sec:tiam}
We use the following set of 24 COCO labels $O =$ \textit{bicycle, car, motorcycle, truck, donut, bench, bird, cat, dog, horse, sheep, cow, elephant, bear, zebra, giraffe, banana, apple, broccoli, carrot, chair, couch, oven, refrigerator}. The templates are :
\begin{itemize}
    \item two entities: ``a photo of $det(o_1)$  $o_1$ and  $det(o_2)$  $o_2$''
    \item three entities: ``a photo of $det(o_1)$  $o_1$'' next to $det(o_2)$  $o_2$ and $det(o_3)$  $o_3$''
\end{itemize}
with $o_i \in O$ and $det(.)$ the correct article depending on the object $o_i$.

With attribute, we retake the same set of objects $O$ with the following set of attributes $\mathcal{A}=\{\mathit{red, green, blue, purple, pink, yellow}\}$. We used the following templates:
\begin{itemize}
    \item two colored entities : ``a photo of $det(a_1)$ $a_1$ $o_1$ and $det(a_2)$ $a_2$ $o_2$ ''
    \item three colored entities : ``a photo of $det(a_1)$ $a_1$ $o_1$, $det(a_2)$ $a_2$ $o_2$ and  $det(a_3)$ $a_3$ $o_3$''
\end{itemize}
with $o_i \in O$, $a_i \in \mathcal{A}$ and $det(.)$ the correct article depending on the attribute $a_i$.

We then generate all the combinations and following \cite{Grimal_2024_WACV}, we can obtain an approximation by sampling 300 prompts and generate 16 images per prompt using the same seeds. We follow the main implementation, we detect the presence of an object with YOLOv8 \citep{yolov8_ultralytics} and accept the presence of the object if confidence $\geq 0.25$. For an image to be considered well-generated, the requested entities must be correctly detected. Additionally, in the case of colored entities, both the entity detection and the color attribution must be accurate.

In comparison with the Attend\&Excite evaluation setup, in the case of attribute binding,  each entity is qualified by an attribute. In \autoref{tab:stats_2_obj} and \autoref{tab:stats_3_obj}, we present the distribution of couple and trio of meta-class of entities following this classification of different labels :
\begin{itemize}
    \item Animal : bird, cat, dog, horse, sheep, cow, elephant, bear, zebra, giraffe.
    \item Objects : bicycle, car, motorcycle, truck, donut, bench, banana, apple, broccoli, carrot, chair, couch, oven, refrigerator.
\end{itemize}
For reproduction of the experiments, we release the datasets\footnote{\url{https://huggingface.co/datasets/Paulgrimal/2_entities}}\footnote{\url{https://huggingface.co/datasets/Paulgrimal/2_colored_entities}}\footnote{\url{https://huggingface.co/datasets/Paulgrimal/3_entities}}\footnote{\url{https://huggingface.co/datasets/Paulgrimal/3_colored_entities}}.

\begin{table*}
    \centering

    \begin{tabular}{llll}
    \toprule
    Dataset & Animal-Animal & Animal-Object & Object-Object \\
    \midrule
    2 entities & 47 & 151 & 102 \\
    2 colored entities & 45 & 140 & 115 \\
    \midrule
    2 entities + 2 colored entities & 92 & 291 & 217 \\
    \bottomrule
\end{tabular}

    \caption{Number of associations of classes in the datasets of prompts with two entities.}
    \label{tab:stats_2_obj}
\end{table*}

\begin{table*}
    \centering

    \begin{tabular}{lllll}
    \toprule
    Dataset & {\scriptsize	Animal-Animal} & {\scriptsize  Animal-Animal} & {\scriptsize	 Animal-Object} & {\scriptsize Object-Object} \\
    & {\scriptsize	Animal} & {\scriptsize  Object} & {\scriptsize	Object} & {\scriptsize Object} \\

    \midrule
    3 entities & 17 & 98 & 135 & 50 \\
    3 colored entities & 12 & 87 & 139 & 62 \\
    \midrule
    3 entities + 3 colored entities & 29 & 185 & 274 & 112 \\
\bottomrule

\end{tabular}

    \caption{Number of associations of classes in the datasets of prompts with three entities.}
    \label{tab:stats_3_obj}
\end{table*}

\clearpage
\subsection{Attend\&Excite evaluation}
\label{sec:eval_attend}
Attend\&Excite uses CLIP\footnote{\url{https://huggingface.co/openai/clip-vit-base-patch16}} \citep{radford2021learning} and BLIP\footnote{\url{https://huggingface.co/Salesforce/blip-image-captioning-base}} \citep{li2022blip} for evaluation. They compute scores using the cosine similarity of CLIP embedding. To have an average semantic embedding to compute, they create 80 derived of the prompt using 80 templates such as:
\begin{itemize}
    \item "a bad photo of a \{\}",
    \item "a photo of many \{\}",
    \item "a sculpture of a \{\}",
\end{itemize}available on their github\footnote{\url{https://github.com/yuval-alaluf/Attend-and-Excite/}}. Then they fill out the \{\} with the entities in the original prompt. After that, they compute the CLIP embedding and average among the 80 created prompts.

We detail how they compute each score:
\begin{itemize}
    \item \textit{Full Prompt Similarity}: Cosine similarity between the CLIP embedding of the generated image and the average embedding from the 80 templates.
    \item \textit{Minimum Object Similarity}: Average text CLIP embedding for each entity is computed from the templates. Cosine similarity between the generated image and each average embedding corresponding to an entity and the minimum similarity is reported.
    \item  \textit{Text-Text Similarity}: The caption of the generated image (with BLIP) is compared with the average embedding of the 80 templates of the original prompt using cosine similarity.
\end{itemize}

\begin{table*}[h]
    \centering

    \begin{tabular}{rccllll}
\toprule
& \iteref &\gsng &Methods & Full Prompt & Minimum Object & Text-Text \\
\midrule
\multirow[c]{7}{*}{\rotatebox{90}{\textbf{SD 1.4}}} & 0 & \xmark & Stable Diffusion & 0.3313$^{\pm 0.0375}$ & 0.2400$^{\pm 0.0377}$ & 0.7682$^{\pm 0.1017}$ \\
\cmidrule(lr){2-7}
& \multirow[c]{2}{*}{1} & \multirow[c]{2}{*}{\xmark} & InitNo & 0.3411$^{\pm 0.0350}$ & 0.2512$^{\pm 0.0328}$ & 0.7901$^{\pm 0.1012}$ \\
& &  & Ours & 0.3470$^{\pm 0.0336}$ & 0.2564$^{\pm 0.0308}$ & 0.7979$^{\pm 0.0990}$ \\
\cmidrule(lr){2-7}
& \multirow[c]{3}{*}{3} & \multirow[c]{3}{*}{\cmark} & Divide\&Bind & 0.3468$^{\pm 0.0295}$ & 0.2597$^{\pm 0.0246}$ & 0.8065$^{\pm 0.0962}$ \\
& &  & Attend\&Excite & 0.3509$^{\pm 0.0296}$ & 0.2634$^{\pm 0.0226}$ & 0.8032$^{\pm 0.0964}$ \\
& &  & InitNO+ & \underline{0.3520}$^{\pm 0.0285}$ & 0.2638$^{\pm 0.0211}$ & 0.8076$^{\pm 0.0951}$ \\
\cmidrule(lr){2-7}
& 0 & \cmark & Syngen & 0.3518$^{\pm 0.0282}$ & \underline{0.2640}$^{\pm 0.0231}$ & \underline{0.8122}$^{\pm 0.0970}$ \\
\cmidrule(lr){2-7}
& 1 & \cmark & Ours+ & \textbf{0.3522}$^{\pm 0.0270}$ & \textbf{0.2643}$^{\pm 0.0213}$ & \textbf{0.8133}$^{\pm 0.0960}$ \\
\midrule
\multirow[c]{2}{*}{\rotatebox{90}{\textbf{SD 3}}} & 0 & \xmark & Stable Diffusion &  0.3529$^{\pm 0.0294}$ & 0.2616$^{\pm 0.0244}$ & 0.8181$^{\pm 0.0921}$ \\
\cmidrule(lr){2-7}
& 1 & \xmark & Ours & \textbf{0.3535}$^{\pm 0.0281}$ & \textbf{0.2619}$^{\pm 0.0226}$ & \textbf{0.8190}$^{\pm 0.0928}$ \\

\bottomrule
\end{tabular}

    \caption{Similarity scores based on \protect\citep{chefer2023attendandexcite} for two entities. The exponents present the standard deviations. Best values are in bold, with second-best underlined for SD 1.4. For SD 3, only best values are in bold.}
    \label{tab:its_2_objects}

\end{table*}

\begin{table*}[h]
    \centering
    \begin{tabular}{rccllll}
\toprule
& \iteref &\gsng &Methods & Full Prompt & Minimum Object & Text-Text \\
\midrule
\multirow[c]{7}{*}{\rotatebox{90}{\textbf{SD 1.4}}} & 0 & \xmark & Stable Diffusion & 0.3450$^{\pm 0.0381}$ & 0.2063$^{\pm 0.0293}$ & 0.7322$^{\pm 0.1012}$ \\
\cmidrule(lr){2-7}
& \multirow[c]{2}{*}{1} & \multirow[c]{2}{*}{\xmark} & InitNo & 0.3528$^{\pm 0.0364}$ & 0.2106$^{\pm 0.0302}$ & 0.7408$^{\pm 0.1041}$ \\
& &  & Ours & 0.3639$^{\pm 0.0356}$ & 0.2204$^{\pm 0.0305}$ & 0.7568$^{\pm 0.1013}$ \\
\cmidrule(lr){2-7}
& \multirow[c]{3}{*}{3} & \multirow[c]{3}{*}{\cmark} & Divide\&Bind & 0.3687$^{\pm 0.0341}$ & 0.2282$^{\pm 0.0281}$ & 0.7618$^{\pm 0.1038}$ \\
& &  & Attend\&Excite & 0.3708$^{\pm 0.0326}$ & 0.2327$^{\pm 0.0252}$ & 0.7582$^{\pm 0.1026}$ \\
& &  & InitNO+ & 0.3719$^{\pm 0.0324}$ & \underline{0.2331}$^{\pm 0.0240}$ & 0.7594$^{\pm 0.1048}$ \\
\cmidrule(lr){2-7}
& 0 & \cmark & Syngen & \underline{0.3750}$^{\pm 0.0311}$ & 0.2320$^{\pm 0.0277}$ & \underline{0.7660}$^{\pm 0.1066}$ \\
\cmidrule(lr){2-7}
& 1 & \cmark & Ours+ & \textbf{0.3772}$^{\pm 0.0299}$ & \textbf{0.2349}$^{\pm 0.0253}$ & \textbf{0.7698}$^{\pm 0.1056}$ \\
\midrule
\multirow[c]{2}{*}{\rotatebox{90}{\textbf{SD 3}}} & 0 & \xmark & Stable Diffusion & 0.3833$^{\pm 0.0309}$ & 0.2346$^{\pm 0.0255}$ & 0.7876$^{\pm 0.0966}$ \\
\cmidrule(lr){2-7}
& 1 & \xmark & Ours & \textbf{0.3863}$^{\pm 0.0281}$ & \textbf{0.2373}$^{\pm 0.0226}$ & \textbf{0.7908}$^{\pm 0.0951}$ \\
\bottomrule
\end{tabular}

    \caption{    Similarity scores based on \protect\citep{chefer2023attendandexcite} for three entities. The exponents present the standard deviations. Best values are in bold, with second-best underlined for SD 1.4. For SD 3, only best values are in bold.}
    \label{tab:its_3_objects}
\end{table*}

\begin{table*}[h]
    \centering
    \begin{tabular}{rccllll}
\toprule
& \iteref &\gsng &Methods & Full Prompt & Minimum Object & Text-Text \\
\midrule
\multirow[c]{7}{*}{\rotatebox{90}{\textbf{SD 1.4}}} & 0 & \xmark & Stable Diffusion & 0.3527$^{\pm 0.0343}$ & 0.2483$^{\pm 0.0393}$ & 0.7208$^{\pm 0.1130}$ \\
\cmidrule(lr){2-7}
& \multirow[c]{2}{*}{1} & \multirow[c]{2}{*}{\xmark} & InitNo & 0.3639$^{\pm 0.0337}$ & 0.2618$^{\pm 0.0363}$ & 0.7329$^{\pm 0.1120}$ \\
&  &  & Ours & 0.3720$^{\pm 0.0330}$ & 0.2699$^{\pm 0.0329}$ & 0.7420$^{\pm 0.1143}$ \\
\cmidrule(lr){2-7}
& \multirow[c]{3}{*}{3} & \multirow[c]{3}{*}{\cmark} & Divide\&Bind & 0.3688$^{\pm 0.0303}$ & 0.2711$^{\pm 0.0297}$ & 0.7317$^{\pm 0.1180}$ \\
& &  & Attend\&Excite & 0.3767$^{\pm 0.0298}$ & 0.2782$^{\pm 0.0267}$ & 0.7422$^{\pm 0.1150}$ \\
& &  & InitNO+ & \textbf{0.3787}$^{\pm 0.0289}$ & \textbf{0.2792}$^{\pm 0.0256}$ & 0.7453$^{\pm 0.1129}$ \\
\cmidrule(lr){2-7}
& 0 & \cmark & Syngen & \underline{0.3784}$^{\pm 0.0309}$ & 0.2774$^{\pm 0.0296}$ & \textbf{0.7534}$^{\pm 0.1175}$ \\
\cmidrule(lr){2-7}
& 1 & \cmark & Ours+ & 0.3780$^{\pm 0.0304}$ & \underline{0.2784}$^{\pm 0.0280}$ & \underline{0.7483}$^{\pm 0.1196}$ \\
\midrule
\multirow[c]{2}{*}{\rotatebox{90}{\textbf{SD 3}}} & 0 & \xmark & Stable Diffusion & 0.3863$^{\pm 0.0273}$ & 0.2806$^{\pm 0.0259}$ & \textbf{0.7731}$^{\pm 0.1225}$ \\
\cmidrule(lr){2-7}
& 1 & \xmark & Ours & \textbf{0.3864}$^{\pm 0.0262}$ & \textbf{0.2812}$^{\pm 0.0241}$ & 0.7708$^{\pm 0.1238}$ \\
\bottomrule

\end{tabular}

    \caption{Similarity scores based on \protect\citep{chefer2023attendandexcite} for two colored entities. The exponents present the standard deviations. Best values are in bold, with second-best underlined for Stable Diffusion 1.4. For Stable Diffusion 3, only best values are in bold.}
    \label{tab:its_2_objects_color}
\end{table*}

\begin{table*}[h]
    \centering
    \begin{tabular}{rccllll}
\toprule
& \iteref &\gsng &Methods & Full Prompt & Minimum Object & Text-Text \\
\midrule
\multirow[c]{7}{*}{\rotatebox{90}{\textbf{SD 1.4}}} & 0 & \xmark & Stable Diffusion & 0.3519$^{\pm 0.0331}$ & 0.2148$^{\pm 0.0297}$ & 0.6505$^{\pm 0.1017}$ \\
\cmidrule(lr){2-7}
& \multirow[c]{2}{*}{1} & \multirow[c]{2}{*}{\xmark} & InitNo & 0.3633$^{\pm 0.0317}$ & 0.2211$^{\pm 0.0305}$ & 0.6578$^{\pm 0.1026}$ \\
&  &  & Ours & 0.3707$^{\pm 0.0313}$ & 0.2274$^{\pm 0.0299}$ & 0.6621$^{\pm 0.1043}$ \\
\cmidrule(lr){2-7}
& \multirow[c]{3}{*}{3} & \multirow[c]{3}{*}{\cmark} & Divide\&Bind & 0.3689$^{\pm 0.0298}$ & 0.2305$^{\pm 0.0279}$ & 0.6542$^{\pm 0.1016}$ \\
&  &  & Attend\&Excite & 0.3772$^{\pm 0.0292}$ & \underline{0.2388}$^{\pm 0.0261}$ & 0.6557$^{\pm 0.1024}$ \\
& &  & InitNO+ & \textbf{0.3809}$^{\pm 0.0297}$ & \textbf{0.2403}$^{\pm 0.0256}$ & 0.6565$^{\pm 0.1048}$ \\
\cmidrule(lr){2-7}
& 0 & \cmark & Syngen & 0.3754$^{\pm 0.0305}$ & 0.2308$^{\pm 0.0294}$ & \textbf{0.6715}$^{\pm 0.1065}$ \\
\cmidrule(lr){2-7}
& 1 & \cmark & Ours+ & \underline{0.3776}$^{\pm 0.0302}$ & 0.2346$^{\pm 0.0290}$ & \underline{0.6673}$^{\pm 0.1065}$ \\
\midrule
\multirow[c]{2}{*}{\rotatebox{90}{\textbf{SD 3}}} & 0 & \xmark & Stable Diffusion & 0.3998$^{\pm 0.0253}$ &0.2460$^{\pm 0.0231}$ & 0.6726$^{\pm 0.1104}$ \\
\cmidrule(lr){2-7}
& 1 & \xmark & Ours & \textbf{0.4025}$^{\pm 0.0242}$ & \textbf{0.2486}$^{\pm 0.0211}$ & \textbf{0.6744}$^{\pm 0.1104}$ \\
\bottomrule
\end{tabular}

    \caption{Similarity scores based on \protect\citep{chefer2023attendandexcite} for three colored entities. The exponents present the standard deviations. Best values are in bold, with second-best underlined for Stable Diffusion 1.4. For Stable Diffusion 3, only best values are in bold.}
    \label{tab:its_3_objects_color}
\end{table*}

In our case, we compute the score for each dataset. In addition to the results presented in the main paper, we provide average evaluations for the all datasets with the standard deviation:

\begin{itemize}
    \item two entities \autoref{tab:its_2_objects},
    \item three entities \autoref{tab:its_3_objects},
    \item two colored entities \autoref{tab:its_2_objects_color},
    \item three colored entities  \autoref{tab:its_3_objects_color}.
\end{itemize}
In the context of one-step refinement, our method consistently outperforms InitNO. With GSN guidance, we observe slight improvements for the three-entities datasets compared to other approaches; however, our scores are lower for datasets that include colors, which may be explained by the limitations of CLIP-based metrics, as they have a bags-of-words behavior\citep{yuksekgonul2023when}: inadequate relational understanding, frequent errors in associating objects with their attributes, and a significant lack of sensitivity to the order of elements. In addition, the close similarity of the scores, along with the large standard deviations, suggests that this evaluation used might not be accurately detecting significant differences between methods. This brings into question whether the results are truly meaningful, highlighting the need for further research to assess the validity and reliability of this metric in evaluating text-image alignment performance.

\clearpage
\subsection{User study}
\label{sec:user_study}

\begin{figure*}
    \centering
    \begin{subfigure}[b]{0.45\textwidth}
        \centering
        \includegraphics[width=\linewidth]{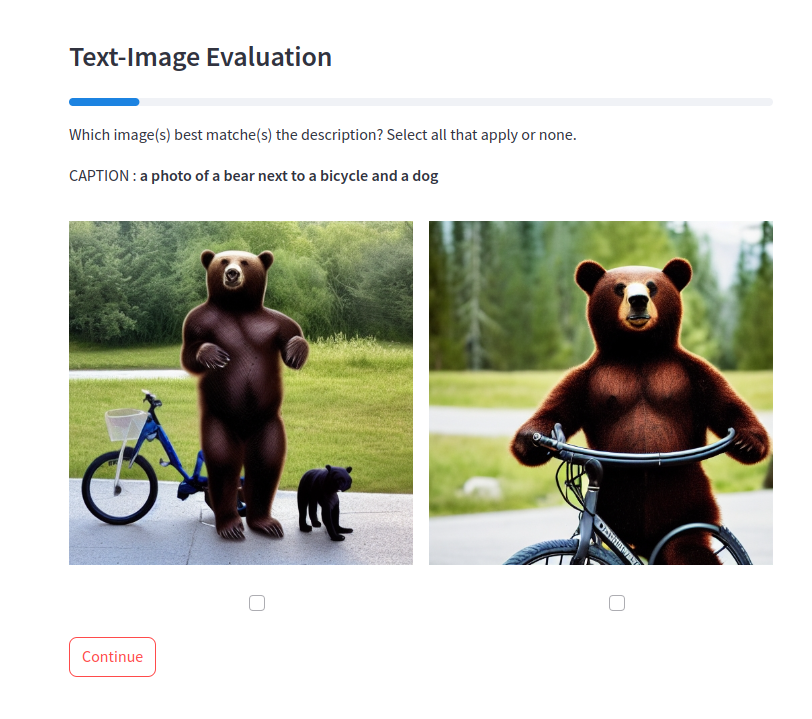}
        \caption{First phase}
    \end{subfigure}
    \hfill
    \begin{subfigure}[b]{0.54\textwidth}
        \centering
        \includegraphics[width=\linewidth]{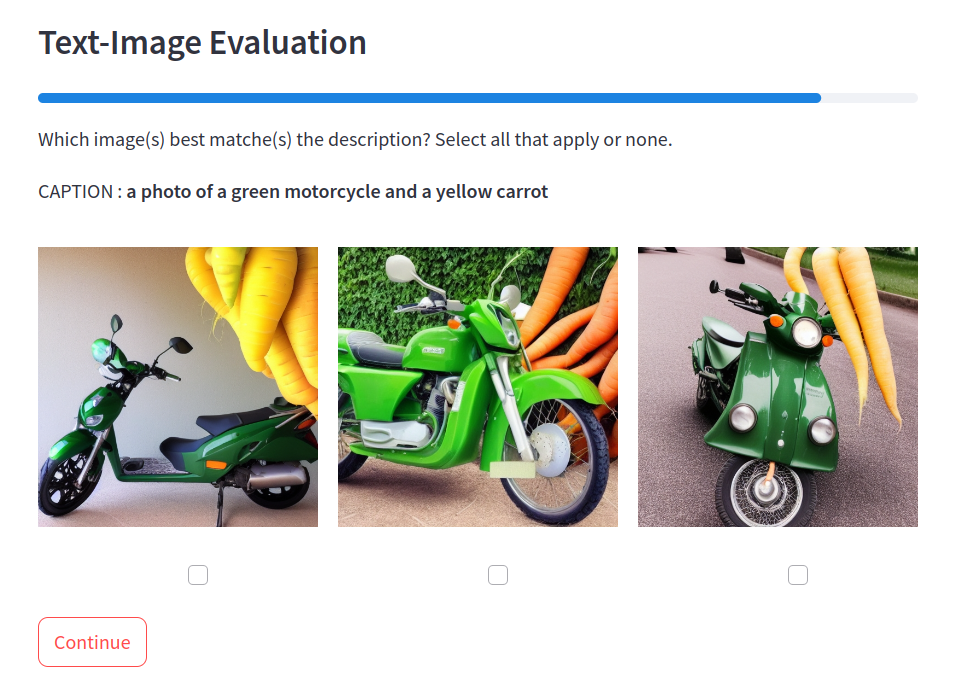}
        \caption{Second phase}
    \end{subfigure}
    \caption{Screenshots of the user study interface.}
    \label{fig:user_study_interface}
\end{figure*}
To compare the methods, we conducted a user study in which participants were shown images generated using the same seed and prompt. The images were randomly sampled from the generated test set. Note that the InitNO method may resample new seeds due to its multi-round iterative refinement step.

Participants were asked to select images that best matched the given prompt. They could choose one, multiple, or none of the images. The study consisted of two phases:

\begin{itemize}
    \item The first phase involved presenting images from Ours and InitNO, representing methods without GSN guidance. Images were shown from the two entities and three entities datasets.
    \item The second phase involved presenting images from Ours+, InitNO+, and Syngen, representing methods with GSN guidance. Images were shown from the two entities, two colored entities, and three entities datasets.
\end{itemize}
The selection of the presented datasets is based on the TIAM score. Without guidance, methods perform too poorly on the two and three colored datasets. With guidance, methods still perform poorly on the three colored dataset.

Each participant was asked to respond to 16 prompts in the first phase and 21 prompts in the second phase. The results from 22 participants, who were shown the same set of images, were used to compute inter-rater reliability using Fleiss' kappa \citep{fleiss1971mns}, where 0.5 indicates fair agreement \citep{Landis77}. \autoref{fig:user_study_interface} shows the interface used by participants to select the images.

\begin{figure}
    \centering
    \includegraphics[width=0.5\textwidth]{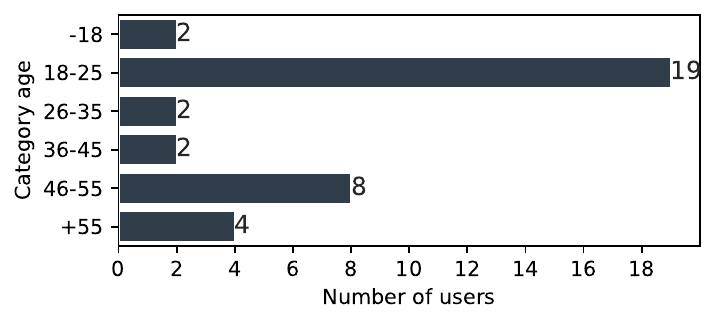}
    \caption{Distribution of participants' ages in the study.}
    \label{fig:distribution_age}
\end{figure}

In total, we had 37 participants, of whom 7 were experts in computer vision. The distribution of participants' age categories is shown in \autoref{fig:distribution_age}.

\clearpage
\subsection{More qualitative samples}
\label{sec:qual_samples}

We provide more examples of generated images with Stable Diffusion 1.4:
\begin{itemize}
    \item without GSN guidance in \autoref{fig:wo_gsng_comparison}, \autoref{fig:wo_qual_1}, \autoref{fig:wo_qual_2}, \autoref{fig:wo_qual_3}, \autoref{fig:wo_qual_4}, \autoref{fig:wo_qual_5},
    \item with GSN guidance in \autoref{fig:w_qual_1}, \autoref{fig:w_qual_2}, \autoref{fig:w_qual_3}, \autoref{fig:w_qual_4}, \autoref{fig:w_qual_5}.
\end{itemize}

We provide examples of generated images with Stable Diffusion 3 in \autoref{fig:sd3_qual_36}, \autoref{fig:sd3_qual_311}, \autoref{fig:sd3_qual_615}, \autoref{fig:sd3_qual_914} and \autoref{fig:sd3_qual_15}.
\newcommand{\captionsdtransformers}{Qualitative comparison between samples generated with SD 3. Images generated with the same set of seeds across the different approaches.}

\newcommand{\captionwogsn}{Qualitative comparison between samples generated with methods without \gsng. Images generated with the same set of seeds across the different approaches, using SD 1.4.}
\newcommand{\captionwgsn}{Qualitative comparison between samples generated with methods with \gsng. Images generated with the same set of seeds across the different approaches, using SD 1.4.}

\begin{figure*}[t]
    \centering
    \input{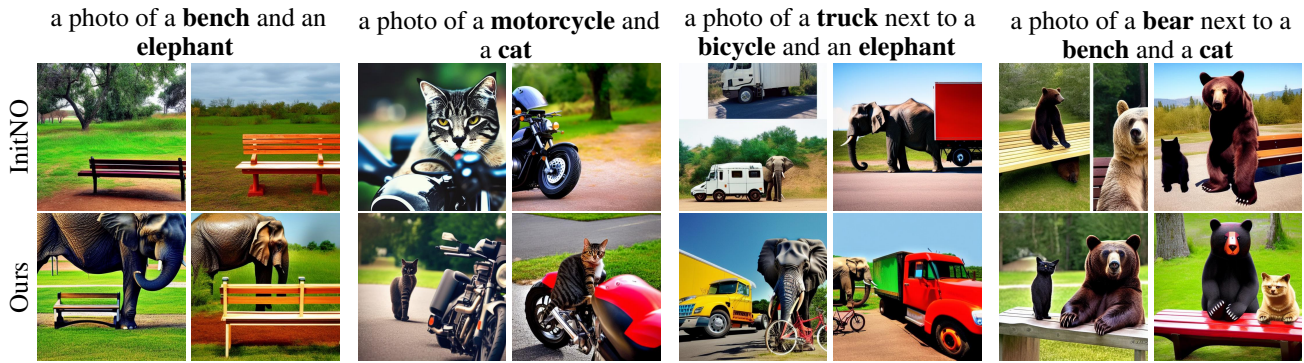}
    \caption{Example of images generated \textit{without} \gsng. (\textit{left} \sdun; \textit{right} \sdtrois). The same seeds were used for all approaches.}
    \label{fig:wo_gsng_comparison}
\end{figure*}

\begin{figure*}[h]
    \centering
    \input{data/illustration/quali_comparison/wo_gsn/5_0}
    \caption{\captionwogsn}
    \label{fig:wo_qual_1}
\end{figure*}

\begin{figure*}[h]
    \centering
    \input{data/illustration/quali_comparison/wo_gsn/5_1}
    \caption{\captionwogsn}
    \label{fig:wo_qual_2}
\end{figure*}

\begin{figure*}[h]
    \centering
    \input{data/illustration/quali_comparison/wo_gsn/11_2}
    \caption{\captionwogsn}
    \label{fig:wo_qual_3}
\end{figure*}

\begin{figure*}[h]
    \centering
    \input{data/illustration/quali_comparison/wo_gsn/11_4}
    \caption{\captionwogsn}
    \label{fig:wo_qual_4}
\end{figure*}

\begin{figure*}[h]
    \centering
    \input{data/illustration/quali_comparison/wo_gsn/14_13}
    \caption{\captionwogsn}
    \label{fig:wo_qual_5}
\end{figure*}

\begin{figure*}[h]
    \centering
    \input{data/illustration/quali_comparison/w_gsn/2_6}
    \caption{\captionwgsn}
    \label{fig:w_qual_1}
\end{figure*}

\begin{figure*}[h]
    \centering
    \input{data/illustration/quali_comparison/w_gsn/2_11}
    \caption{\captionwgsn}
    \label{fig:w_qual_2}
\end{figure*}

\begin{figure*}[h]
    \centering
    \input{data/illustration/quali_comparison/w_gsn/7_10}
    \caption{\captionwgsn}
    \label{fig:w_qual_3}
\end{figure*}

\begin{figure*}[h]
    \centering
    \input{data/illustration/quali_comparison/w_gsn/12_0}
    \caption{\captionwgsn}
    \label{fig:w_qual_4}
\end{figure*}

\begin{figure*}[h]
    \centering
    \input{data/illustration/quali_comparison/w_gsn/15_6}
    \caption{\captionwgsn}
    \label{fig:w_qual_5}
\end{figure*}

\begin{figure*}[h]
    \centering
    \input{data/illustration/quali_comparison/sd3/3_6}
    \caption{\captionsdtransformers}
    \label{fig:sd3_qual_36}
\end{figure*}

\begin{figure*}[h]
    \centering
    \input{data/illustration/quali_comparison/sd3/3_11}
    \caption{\captionsdtransformers}
    \label{fig:sd3_qual_311}
\end{figure*}

\begin{figure*}[h]
    \centering
    \input{data/illustration/quali_comparison/sd3/6_15}
    \caption{\captionsdtransformers}
    \label{fig:sd3_qual_615}
\end{figure*}

\begin{figure*}[h]
    \centering
    \input{data/illustration/quali_comparison/sd3/9_14}
    \caption{\captionsdtransformers}
    \label{fig:sd3_qual_914}
\end{figure*}

\begin{figure*}[h]
    \centering
    \input{data/illustration/quali_comparison/sd3/15_11}
    \caption{\captionsdtransformers}
    \label{fig:sd3_qual_15}
\end{figure*}

\clearpage
\subsection{Reporting the scores values and additional results}
\label{sec:results}
\subsubsection{Validation set}
\label{sec:validation}

\begin{figure*}[h]
    \centering
    \includegraphics[width=1\linewidth]{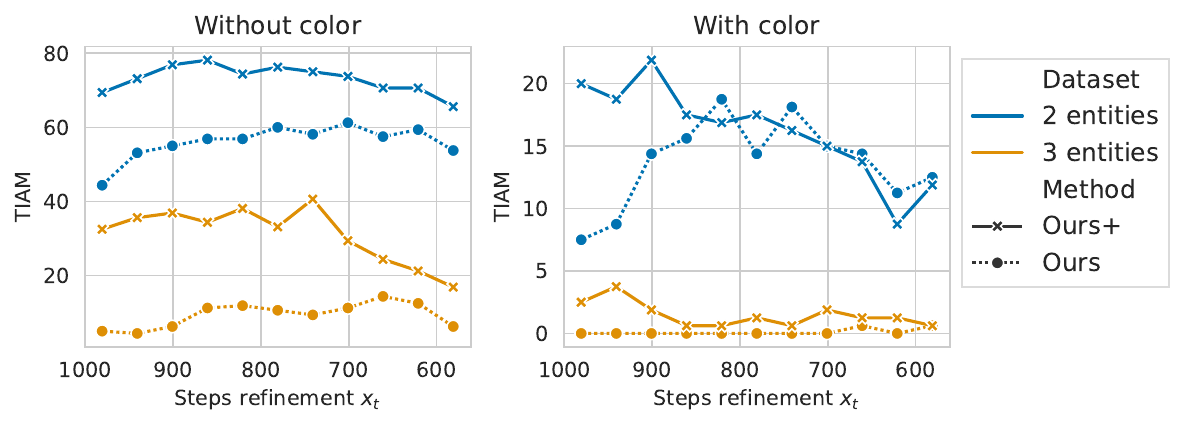}
    \caption{TIAM score of datasets of 10 prompts with 2 and 3 objects as a function of the refinement step used. On the right, entities are bound with colors, we then use the TIAM score with color ground truth.}
    \label{fig:prompt_val_lineplot}
\end{figure*}

\begin{figure*}
    \begin{minipage}[t]{0.49\textwidth}
        \centering
        \includegraphics[width=1\linewidth]{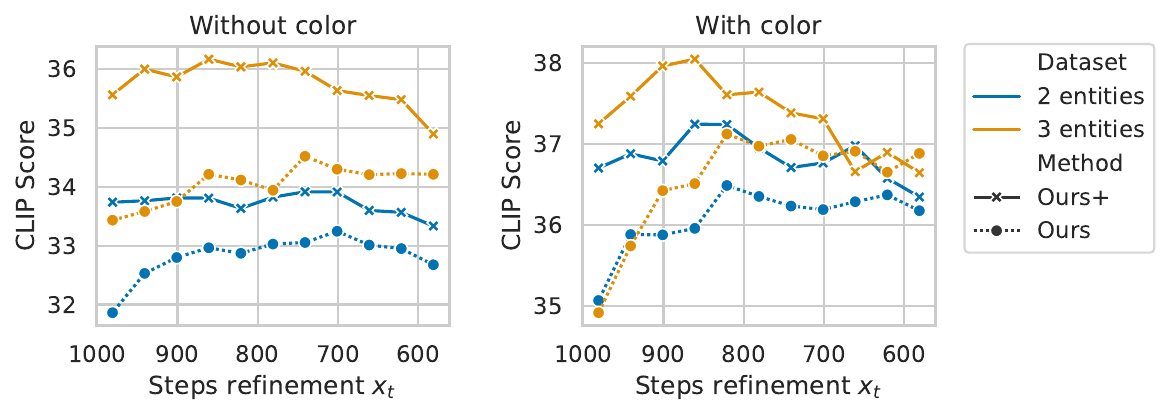}
        \caption{CLIP score according to the iterative refinement step used for the validation datasets.}
        \label{fig:val_clip_score}
    \end{minipage}
    \hfill
    \begin{minipage}[t]{0.49\textwidth}
        \centering
        \includegraphics[width=1\linewidth]{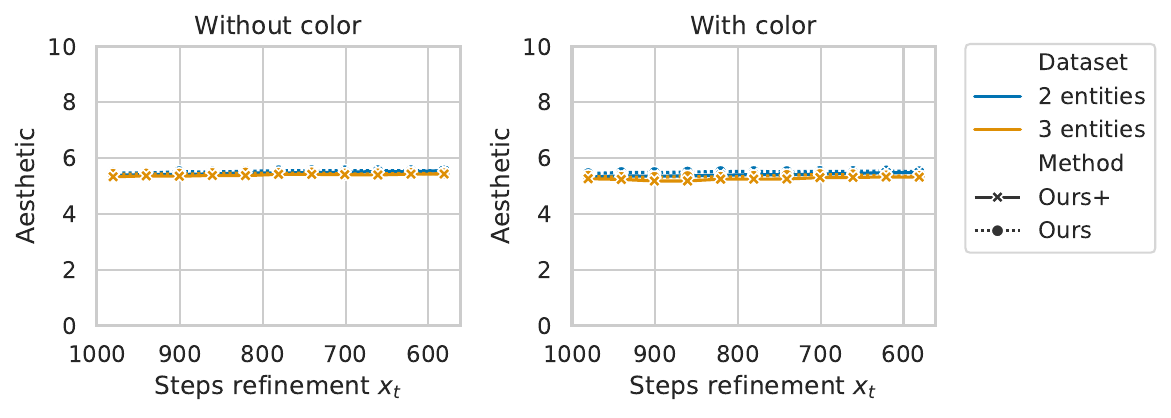}
        \caption{Aesthetic score according to the iterative refinement step used for the validation. The Aesthetic score is between 1 and 10.}
        \label{fig:val_aesthetic}
    \end{minipage}

\end{figure*}

We report the TIAM scores on the validation dataset in \autoref{tab:prompt_val} and we represent the scores as a function of refinement steps used in \autoref{fig:prompt_val_lineplot}. Additionally, we present the CLIP Score (in \autoref{fig:val_clip_score}) and Aesthetic score (\autoref{fig:val_aesthetic}) according to the refinements steps used.

\begin{table*}
    \centering
    \caption{TIAM score according to the steps used for the refinement with datasets with 10 prompts (when color TIAM score ground truth colors that is displayed).}
    \begin{tabular}{rcccccccc}
\toprule
\multirow[c]{3}{*}{\rotatebox{0}{\shortstack{step of\\iterative\\refinement}}} & \multicolumn{4}{c}{2 entities} & \multicolumn{4}{c}{3 entities} \\
 & \multicolumn{2}{c}{wo color} & \multicolumn{2}{c}{color} & \multicolumn{2}{c}{wo color} & \multicolumn{2}{c}{color} \\
 & \o & GSNg & \o & GSNg & \o & GSNg & \o & GSNg \\
\midrule
981 & 44.38 & 69.38 & 7.50 & 20.00 & 5.00 & 32.50 & 0.00 & 2.50 \\
941 & 53.12 & 73.12 & 8.75 & 18.75 & 4.38 & 35.62 & 0.00 & 3.75 \\
901 & 55.00 & 76.88 & 14.38 & 21.88 & 6.25 & 36.88 & 0.00 & 1.88 \\
861 & 56.88 & 78.13 & 15.62 & 17.50 & 11.25 & 34.38 & 0.00 & 0.62 \\
821 & 56.88 & 74.38 & 18.75 & 16.88 & 11.88 & 38.12 & 0.00 & 0.62 \\
781 & 60.00 & 76.25 & 14.38 & 17.50 & 10.63 & 33.13 & 0.00 & 1.25 \\
741 & 58.13 & 75.00 & 18.13 & 16.25 & 9.38 & 40.63 & 0.00 & 0.62 \\
701 & 61.25 & 73.75 & 15.00 & 15.00 & 11.25 & 29.38 & 0.00 & 1.88 \\
661 & 57.50 & 70.62 & 14.38 & 13.75 & 14.38 & 24.38 & 0.62 & 1.25 \\
621 & 59.38 & 70.62 & 11.25 & 8.75 & 12.50 & 21.25 & 0.00 & 1.25 \\
581 & 53.75 & 65.62 & 12.50 & 11.88 & 6.25 & 16.88 & 0.62 & 0.62 \\
\bottomrule
\end{tabular}

    \label{tab:prompt_val}
\end{table*}

\subsubsection{Test set}
\label{sec:test}

\begin{figure*}[h]
    \begin{minipage}[t]{0.49\textwidth}
        \centering
        \includegraphics[width=1\linewidth]{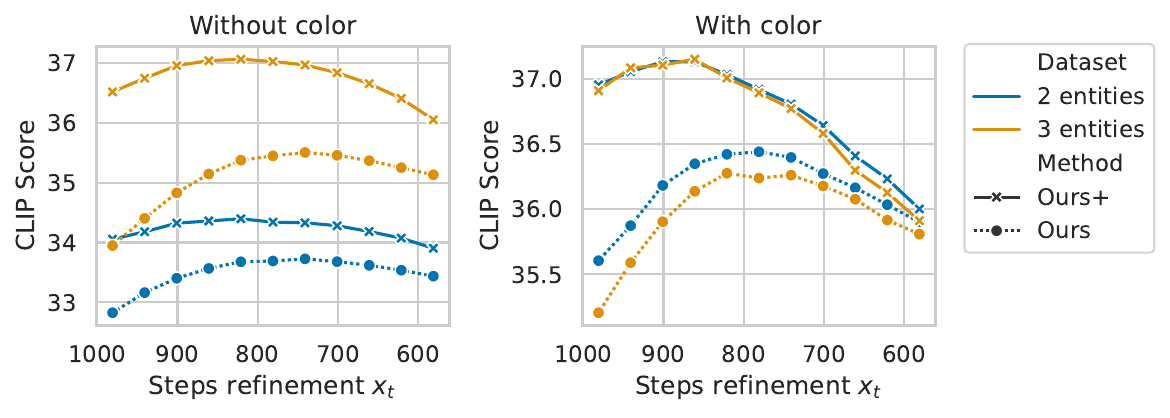}
        \caption{CLIP score according to the iterative refinement step used for the test datasets.}
        \label{fig:test_clip}
    \end{minipage}
    \hfill
    \begin{minipage}[t]{0.49\textwidth}
        \centering
        \includegraphics[width=1\linewidth]{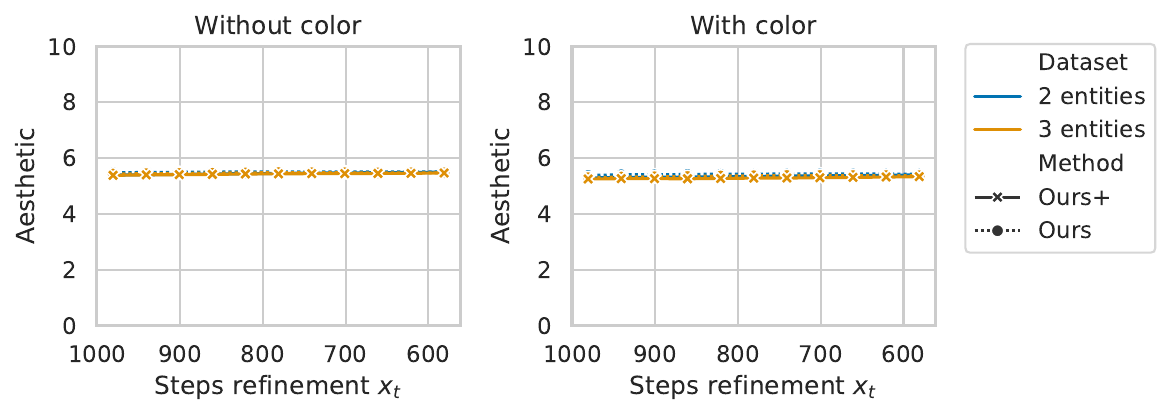}
        \caption{Aesthetic score according to the iterative refinement step used for the validation. The Aesthetic score is between 1 and 10.}
        \label{fig:test_aesthetic}
    \end{minipage}
\end{figure*}

For our methods, we report the TIAM score according to the iterative refinement steps used for the line plot in the main paper, as shown in \autoref{tab:recap_score_steps}. Additionally, we present the CLIP Score (\autoref{fig:test_clip}) and Aesthetic Score (\autoref{fig:test_aesthetic}) corresponding to the refinement steps applied in our methods. Notably, we observe that the Aesthetic Score remains constant regardless of the iterative refinement steps used. Furthermore, we observe similar trends to those reported in the main paper regarding the TIAM score. Specifically, applying iterative refinement at slightly later diffusion steps appears to improve the CLIP score. However, delaying the refinement too much results in a decline in performance over time.

\begin{table*}
    \centering
    \caption{TIAM score as a function of different iterative refinement steps.}
    \begin{tabular}{rcccccccc}
\toprule

\multirow[c]{3}{*}{\rotatebox{0}{\shortstack{step of\\iterative\\refinement}}} & \multicolumn{4}{c}{2 entities} & \multicolumn{4}{c}{3 entities} \\
 & \multicolumn{2}{c}{w/o colors} & \multicolumn{2}{c}{colors} & \multicolumn{2}{c}{w/o colors} & \multicolumn{2}{c}{colors} \\
 & \o & GSNg & \o & GSNg & \o & GSNg & \o & GSNg \\
\midrule
981 & 58.77 & 78.83 & 7.81 & 19.46 & 13.25 & 43.02 & 0.33 & 2.85 \\
941 & 62.00 & 81.10 & 8.42 & 20.54 & 17.08 & 45.79 & 0.29 & 2.77 \\
901 & 65.10 & 81.46 & 9.67 & 20.08 & 20.29 & 47.69 & 0.40 & 2.29 \\
861 & 65.77 & 81.02 & 9.13 & 19.46 & 22.02 & 49.52 & 0.38 & 2.48 \\
821 & 65.81 & 80.56 & 8.71 & 18.21 & 23.12 & 48.98 & 0.38 & 2.25 \\
781 & 66.56 & 79.10 & 8.42 & 16.48 & 25.21 & 48.65 & 0.38 & 1.88 \\
741 & 66.42 & 78.25 & 8.38 & 15.17 & 25.40 & 47.50 & 0.38 & 1.52 \\
701 & 65.38 & 77.23 & 8.35 & 12.83 & 24.54 & 44.35 & 0.46 & 1.35 \\
661 & 64.77 & 74.40 & 7.60 & 11.58 & 23.67 & 41.10 & 0.31 & 0.88 \\
621 & 63.31 & 71.60 & 7.21 & 10.19 & 23.17 & 38.15 & 0.38 & 0.54 \\
581 & 61.81 & 69.62 & 6.71 & 8.62 & 22.19 & 33.96 & 0.31 & 0.58 \\
\bottomrule
\end{tabular}

    \label{tab:recap_score_steps}
\end{table*}

\begin{figure*}[h]
    \centering
    \input{data/fig/fig_concat/boxplots}
    \caption{TIAM aggregate per seed for the 16 seeds  per dataset. “+” shows the mean.}
    \label{fig:boxplot_tiam}
\end{figure*}

We aggregate the TIAM scores per seed across all datasets and methods, with the results shown in \autoref{fig:boxplot_tiam}. The accuracy of InitNO and InitNO+ is somewhat inflated due to their multi-round optimization process, where they resample noise if the initial seed does not perform well, leading to artificially improved results. Our best setup, Ours+, consistently achieves higher average scores than our closest competitor, Syngen. We observe greater robustness across seeds, reflected in a lower interquartile range across all datasets, indicating a higher success rate.

\begin{figure*}[h]
    \centering
    \input{data/fig/fig_concat/proportion_tiam}
    \caption{The proportion of occurrences for each entity based on its position within the prompt across all datasets. Here, we focus solely on the detection of entities, regardless of whether their colors are incorrectly attributed.}
    \label{fig:proportion_tiam}
\end{figure*}
Grimal~\citet{Grimal_2024_WACV} observed that entities positioned earlier in a prompt tend to appear more frequently than those listed later. In \autoref{fig:proportion_tiam}, we report the proportion of occurrences of entities based on their position in the prompt. This trend persists across most methods, with the exception of Ours+ and Syngen, particularly for prompts involving two or three colored entities, where this bias is less pronounced.

\end{document}